\documentclass{article}

\PassOptionsToPackage{compress}{natbib}
\usepackage[final]{corl_2026}

\usepackage[utf8]{inputenc}
\usepackage[T1]{fontenc}
\usepackage{microtype}
\usepackage{graphicx}
\usepackage{subcaption}
\usepackage{hyperref}
\usepackage{url}
\usepackage{booktabs}
\usepackage{amsmath}
\usepackage{amsfonts}
\usepackage{amssymb}
\usepackage{mathtools}
\usepackage{amsthm}
\usepackage[dvipsnames,table]{xcolor}
\usepackage{float}
\usepackage{wrapfig}
\usepackage{caption}
\usepackage{tabularx}
\usepackage{multirow}
\usepackage{makecell}
\usepackage{colortbl}
\usepackage{enumitem}
\usepackage{xspace}
\usepackage{lipsum}
\usepackage{mathrsfs}
\usepackage{dsfont}
\usepackage{bbm}
\usepackage{listings}
\usepackage{pifont}
\usepackage{siunitx}
\usepackage{duckuments}
\usepackage{tcolorbox}
\tcbuselibrary{listings, breakable}
\usepackage[capitalize,noabbrev]{cleveref}
\usepackage[textsize=tiny]{todonotes}

\newcommand{\ie}{\textit{i}.\textit{e}., }
\newcommand{\eg}{\textit{e}.\textit{g}., }

\newcommand{\Methodname}{RoboCurate\xspace}

\newcommand{\placeholder}[1]{{\color{lightgray}\lipsum[1]}}
\newcommand{\cmark}{\ding{51}}
\newcommand{\xmark}{\ding{55}}

\definecolor{RowHighlight}{gray}{0.9}
\definecolor{pinegreen}{rgb}{0.0, 0.47, 0.44}
\definecolor{cornellred}{rgb}{0.7, 0.11, 0.11}
\definecolor{cadmiumgreen}{rgb}{0.0, 0.42, 0.24}
\definecolor{spirodiscoball}{rgb}{0.06, 0.75, 0.99}
\definecolor{Red7}{rgb}{0.941, 0.243, 0.243}
\definecolor{Eqpink}{RGB}{241,241,214}
\definecolor{Green7}{RGB}{55, 178, 77}
\definecolor{aliceblue}{rgb}{0.91, 0.94, 0.97}
\definecolor{darkblue}{rgb}{0.83, 0.89, 0.97}
\definecolor{SJViolet}{RGB}{105,100,171}
\definecolor{SJRed}{RGB}{237,109,107}
\definecolor{tablegreen}{rgb}{0.91, 0.94, 0.97}
\definecolor{sk}{rgb}{0, 0, 1}

\hypersetup{
  colorlinks=true,
  linkcolor=cornellred,
  citecolor=cadmiumgreen,
  urlcolor=cornellred,
  pdfauthor={Seungku Kim, Suhyeok Jang, Byungjun Yoon, Dongyoung Kim, John Won, Jinwoo Shin}
}

\newtcblisting{promptbox}[2][]{
  title={\textbf{#2}},
  colback=gray!5, colframe=gray!50, fonttitle=\bfseries,
  breakable,                                               
  listing only,
  listing options={breaklines=true, basicstyle=\ttfamily\tiny},
  #1
}

\theoremstyle{plain}

\theoremstyle{definition}

\theoremstyle{remark}

\title{RoboCurate: Harnessing Diversity with Action-Verified Synthetic Data for Robot Learning}

\author{%
    \makebox[0pt][c]{Seungku Kim$^{1*}$\,
    Suhyeok Jang$^{1,2*}$\,
    Byungjun Yoon$^{1}$\,
    Dongyoung Kim$^{1,2}$\,
    John Won$^{1}$\,
    Jinwoo Shin$^{1,2}$}\\
    $^{1}$KAIST\quad
    $^{2}$RLWRLD\\
    \texttt{\{seungkukim,jasuhe900,jinwoos\}@kaist.ac.kr}
}

\begin{document}
\maketitle

\begingroup
\renewcommand{\thefootnote}{}
\footnotetext{$^*$ Equal contribution.\hfill {\textbf{Project page}: \url{https://seungkukim.github.io/robocurate/}}}
\endgroup
\setcounter{footnote}{0}  %
\vspace{-0.1in}

\begin{abstract}
Synthetic data generated by video generative models has shown promise for robot learning as a scalable pipeline, but it often suffers from noisy action labels due to imperfect video generation or inaccurate action prediction. Recently, vision-language models (VLMs) have been leveraged to validate video quality, but they cannot directly evaluate the generated actions themselves. To tackle this issue, we introduce \Methodname,  a novel framework for action-verified synthetic robot data that addresses inconsistencies between generated videos and annotated actions. Specifically, \Methodname executes the actions predicted by an inverse dynamics model (IDM) in a simulator and verifies action consistency by measuring the motion agreement between the simulator rollout and the generated video. To better leverage such a framework, we further expand the scene diversity of synthetic data beyond the available dataset through off-the-shelf generative models with structured editing axes. We then use a VLM to pre-filter physically implausible generated videos before verifying their action labels. Across diverse manipulation scenarios, \Methodname yields substantial relative improvements in success rates compared to using only real data, achieving +90.3\% on GR-1 Tabletop (300 demos), +29.1\% on DexMimicGen in the pre-training setup on simulation benchmarks, and  +179.9\% {and +42.9\%} in the co-finetuning setup on challenging real-world {ALLEX and OpenArm} humanoid benchmarks, respectively.

\end{abstract}

\section{Introduction}
\label{sec:intro}

\begin{figure}[t]
    \vspace{-0.1in}
    \includegraphics[width=\textwidth]{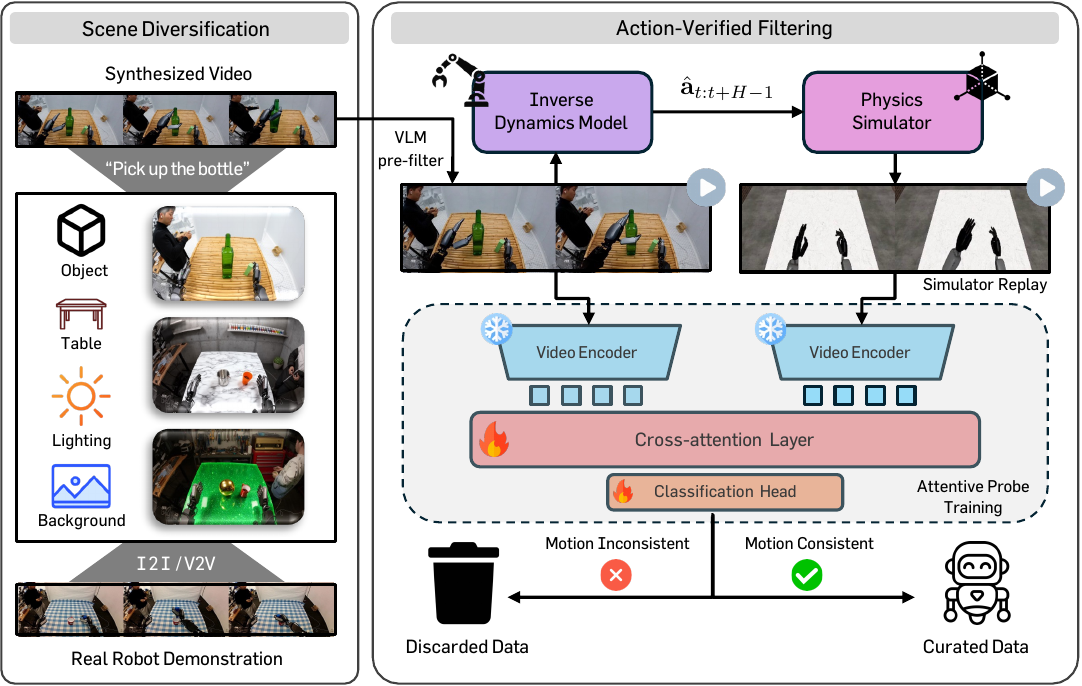}
    \caption{\textbf{Overview of \Methodname}. Our action-verified filtering curates synthetic data by pre-filtering physically implausible videos with a VLM and then verifying motion consistency between a generated video and a simulator replay, discarding samples with motion-inconsistent IDM-predicted actions. To improve filtering effectiveness, we diversify synthetic data by editing four scene components with off-the-shelf I2I and V2V models.}
    \label{figure:overview}
    \vspace{-0.3in}
\end{figure}

Robot foundation models (RFMs; \citep{kim2024openvla,zitkovich2023rt, nvidia2025gr00t,pi0_7_2026}) have achieved strong performance across diverse robotics domains, \eg manipulation, locomotion, and navigation \citep{hirose2025omnivla,zhang2025flexvln}. The key components of this success are large-scale robotics datasets \citep{o2024open,bu2025agibot} spanning diverse tasks and embodiments, and foundation models such as vision-language models (VLMs) and video generative models \citep{cheang2024gr,shen2025videovla}, which support action generation by bridging world knowledge and commonsense with robotic actions. Together, these components serve as the basis for pre-training RFMs on robot demonstration data, enabling them to adapt to downstream tasks with little or no task-specific fine-tuning.

Despite its success, robotics data remains far more limited than vision and language data due to costly and labor-intensive collection processes \citep{khazatsky2024droid, walke2023bridgedata}, which fundamentally constrain large-scale pre-training.
To address this problem, prior research has focused on simulated data as a substitute for real data, but simulation suffers from visual discrepancies with real-world environments and limited knowledge transfer to real-world settings (\ie sim-to-real gap~\citep{zhao2020sim}); it also requires substantial engineering effort to generate diverse environments and action data. Meanwhile, synthetic data generated by video generative models has emerged as a promising alternative, as it is visually closer to real-world data than simulation. Moreover, these models can generate videos of diverse tasks conditioned {on} text prompts \citep{zhou2024robodreamer, fu2025learning}, which can then be annotated with actions through inverse dynamics models (IDMs)~\citep{baker2022video} trained on real robot demonstrations.

Several studies have shown that synthetic data can improve generalization and enable learning novel actions beyond the training dataset \citep{jang2025dreamgen, team2025gigabrain}. However, synthetic data generation pipelines often face quality issues. {Unlike in simulation, the video generation stage can fail to follow input instructions or produce physically implausible videos \citep{bansal2024videophy} (\eg objects overlapping or deforming unnaturally), which leads to incorrect action annotations.} {Moreover, even when the generated videos are realistic, IDMs may predict actions that are inconsistent with the video, which degrade data quality and ultimately lead to suboptimal downstream policy performance.} 
To address this issue, recent work has employed VLMs to judge instruction following or video plausibility~\citep{bansal2024videophy,motamed2025travl}. 
However, such video-level judgments do not verify whether the annotated actions are consistent with the motion shown in the generated video, leaving the resulting synthetic data only partially validated.

\textbf{Contributions.} {In this paper, we propose \Methodname, a novel framework that filters out low-quality synthetic data via simulator-replay consistency, while supporting diverse scene generation.} {The core idea is to leverage a simulator \citep{todorov2012mujoco, zhu2020robosuite} to replay the generated action sequences, producing proxy robot videos with guaranteed correspondence to those actions.} By measuring visual motion consistency between these simulator-replayed videos and the synthesized videos, we can verify whether the generated actions align with the global motion shown in the synthesized videos. Concretely, we replay IDM-predicted actions in a simulator to render a rollout video, and reformulate video-action alignment as a motion-consistency check between the generated video and the simulator replay. To assess this consistency, we train a lightweight attentive probe on top of a pre-trained video encoder~\citep{assran2025v} to classify whether a video pair shares consistent motion patterns and robot geometry. We deliberately construct negative pairs for probe training from real demonstrations to capture motion inconsistencies, eliminating the need for human labels.
To build a diverse synthetic data pool for action-verification, we introduce a scene diversification pipeline based on structured editing axes. Specifically, we employ off-the-shelf generative models~\citep{flux-2-2025, nvidia2025cosmospredict2} to perform image-to-image (I2I) editing or video-to-video (V2V) transfer, diversifying disentangled scene components while preserving manipulation-plausible structure. We then use a VLM to pre-filter generated videos based on the plausibility of robot arm and end-effector trajectories with respect to depth and spatial orientation. Overall, \Methodname yields high-quality synthetic data that reflects a wide range of realistic environmental variations. 

We validate \Methodname across diverse simulation and real-world benchmarks for vision-language-action models (VLAs). In the pre-training setup, our method achieves substantial gains over the real-data-only baseline, including +90.3\% on GR-1 Tabletop~\citep{nvidia2025gr00tn1openfoundation} {(300 demos)} and +29.1\% on DexMimicGen~\citep{jiang2025dexmimicgen}. It also significantly outperforms DreamGen~\citep{jang2025dreamgen}, which lacks our filtering and diversification pipeline. We further observe that these trends generalize to the real ALLEX humanoid, where \Methodname achieves a +179.9\% relative improvement over the real-data-only baseline {on average across tasks, along with emergent success on a novel behavior task (0.0\% $\rightarrow$  25.0\%), and to the real OpenArm humanoid, where it achieves a +42.9\% relative improvement on average and more than doubles the success rate on a novel object task (6.3\% $\rightarrow$ 15.6\%).}

\section{Preliminaries}
\label{sec:prelim}

\textbf{Imitation learning for policy training.}
Our goal is to learn a policy $\pi_\theta$ for robot manipulation from demonstration data. We follow an imitation learning (IL) framework~\citep{zhao2023learning}, where the policy predicts an action chunk $\mathbf{a}_{t:t+H-1}$ conditioned on observation $\mathbf{o}_t$, task instruction $I$, and proprioceptive state $\mathbf{q}_t$. We assume access to a limited real-world dataset $\mathcal{D}_{\text{real}}$ and synthetic data $\mathcal{D}_{\text{syn}}$, and train the policy on the combined dataset
$\mathcal{D} = \mathcal{D}_{\text{real}} \cup \mathcal{D}_{\text{syn}}$. In our experiments, we use a VLA with a diffusion-based action head as the base policy, and optimize it with the following flow-matching objective~\citep{lipman2022flow}:
\begin{equation*}
\mathcal{L}_{\text{IL}}(\theta; \mathcal{D}) =
\mathbb{E}_{t, \mathbf{a}_{t:t+H-1}, \boldsymbol{\epsilon}, \tau}
\Big[
\big\|
\mathbf{v}_\theta(\mathbf{a}_{t:t+H-1}^\tau, \tau \mid \mathbf{o}_t, I, \mathbf{q}_t)
-
(\boldsymbol{\epsilon} - \mathbf{a}_{t:t+H-1})
\big\|_2^2
\Big],
\end{equation*}
where $\tau \in [0,1]$ is sampled from a Beta distribution,
$\boldsymbol{\epsilon} \sim \mathcal{N}(\mathbf{0}, \mathbf{I})$, and
$\mathbf{a}_{t:t+H-1}^{\tau} = (1-\tau) \mathbf{a}_{t:t+H-1} + \tau\boldsymbol{\epsilon}$.
{We note that} $\mathcal{D}_{\text{syn}}$ has zero-padded proprioceptive states, since IDMs do not predict state information.
\section{Method}
\label{sec:method}

{We present \Methodname, a framework for action-verified synthetic robot data that discards low-quality samples while diversifying manipulation scenes along structured editing axes. In~\Cref{subsec:action-verification}, we propose our action-verified filtering strategy that checks, through simulator replay, the consistency of IDM-predicted actions with generated videos that pass a VLM pre-filter, and further show that the same probe can serve as a critic for Best-of-N sampling at generation time. In~\Cref{subsec:generate}, we introduce our diversification pipeline that systematically expands scene diversity through image and video transformations.} We provide an overview of \Methodname in~\Cref{figure:overview}.

\subsection{{Action-Verified Filtering via Simulator Replay}}
\label{subsec:action-verification}

\textbf{Problem description and overview.} {Synthetic robot data produced from generative models still suffers from misaligned video-action pairs that harm policy learning due to imperfect generation. Prior work~\citep{jang2025dreamgen, bansal2024videophy} uses VLMs for video plausibility checks, but this does not check whether the actions are consistent with the video. Verifying this consistency is challenging because directly comparing consistency across modalities (video and action) is non-trivial. To address this, we leverage a physics simulator~\citep{todorov2012mujoco, zhu2020robosuite} to render the IDM-predicted actions as a motion video, enabling video-to-video comparison to filter out inconsistent samples.} 

We represent each synthetic data sample as a pair $(\mathbf{x}_{\text{gen}}, \mathbf{A}_{\text{IDM}})$, where $\mathbf{x}_{\text{gen}}$ denotes a generated video of length $T$ and $\mathbf{A}_{\text{IDM}} = \hat{\mathbf{a}}_{1:T-1}$ denotes its IDM-predicted action sequence (see \cref{appendix:idm} for details). To validate whether $\mathbf{A}_{\text{IDM}}$ is consistent with $\mathbf{x}_{\text{gen}}$, \Methodname{} replays $\mathbf{A}_{\text{IDM}}$ in a physics simulator $f$, and renders the corresponding rollout $\hat{\mathbf{{x}}}_{\text{sim}}=f(\mathbf{A}_{\text{IDM}}, \tilde{\mathbf{o}})$, where $\tilde{\mathbf{o}}$ is an initial observation determined by the simulator and the underlying embodiment (\eg GR-1 humanoid~\citep{nvidia2025gr00tn1openfoundation}). We treat $\hat{\mathbf{{x}}}_{\text{sim}}$ as a rendering of the robot motion specified by $\mathbf{A}_{\text{IDM}}$, thereby reframing action-verification as a motion-consistency comparison between $\mathbf{x}_{\text{gen}}$ and $\hat{\mathbf{{x}}}_{\text{sim}}$. {To this end, we propose to leverage the rich motion prior of a pre-trained video encoder~\citep{assran2025v} and build an attentive probe that classifies motion consistency between the two video embeddings.}

\begin{figure}[t!]
    \vspace{-0.1in}
    \centering
    \includegraphics[width=0.6\textwidth]{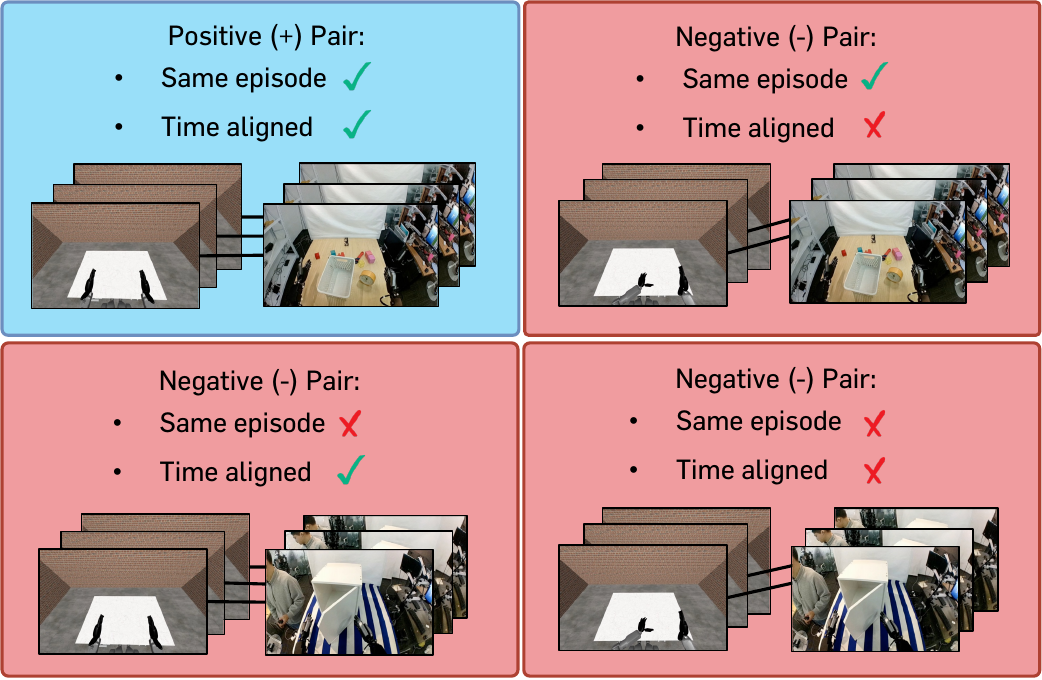}
    \caption{
    \textbf{Examples of training data for the attentive probe.} We construct positive pairs by aligning real-world demonstrations with their simulator replays, while negative pairs are constructed via temporally shifting (intra-episode) and cross-episode swapping within the same skill category.} 
    \label{fig:probe_train_data}
    \vspace{-0.2in}
\end{figure}

\textbf{Probe training data construction.}
 To train the attentive probe for synthetic data filtering, we construct positive and negative video pairs from real-world demonstrations (see~\Cref{fig:probe_train_data}). Specifically, we use a real-world dataset $\mathcal{T}=\{(\mathbf{x}_{i}^{\text{real}}, \mathbf{A}_{i}^{\text{real}})\}_{i=1}^{N_D}$, where $\mathbf{x}_{i}^{\text{real}}$ denotes a real robot video and $\mathbf{A}_{i}^{\text{real}}$ denotes the corresponding ground-truth action sequence. We replay $\mathbf{A}_{i}^{\text{real}}$ in the simulator to obtain the corresponding rollout video
$
\mathbf{x}_{i}^{\text{sim}}
=
f(\mathbf{A}_{i}^{\text{real}}, \tilde{\mathbf{o}})
.
$
Using these paired videos, we construct positive (aligned) pairs
$
\mathcal{P}^{+}
=
\{
(
\mathbf{x}_{i,t}^{\text{real}},
\mathbf{x}_{i,t}^{\text{sim}}
)
\}_{i,t}
,
$
where $\mathbf{x}_{i,t}$ denotes a video clip starting at timestep $t$. We exclude synthetic data from probe training, as its noisy action labels could provide misleading supervision.

To capture motion mismatches at different granularities, we design negative pairs $\mathcal{P}^{-} = \mathcal{P}^{-}_{\text{shift}} \cup \mathcal{P}^{-}_{\text{cross}} \cup \mathcal{P}^{-}_{\text{shift+cross}}$ via two complementary misalignment operations applied individually or jointly: (1) \emph{temporally shifted negatives}, which capture intra-episode variation by pairing clips from different temporal segments within the same episode, and (2) \emph{cross-episode negatives}, which capture inter-episode variation by pairing clips from different episodes within the same skill category (\eg pick-and-place, pour/scoop): 
\begin{equation*}
\begin{split}
\mathcal{P}^{-}_{\text{shift}} = \big\{(\mathbf{x}^{\text{real}}_{i,\,t},\ & \mathbf{x}^{\text{sim}}_{i,\,t'}) \mid t \neq t' \big\},
 \quad \mathcal{P}^{-}_{\text{cross}} = \big\{(\mathbf{x}^{\text{real}}_{i,\,t},\ \mathbf{x}^{\text{sim}}_{j,\,t}) \mid i \neq j \big\}, \\
\mathcal{P}^{-}_{\text{shift+cross}} &= \big\{(\mathbf{x}^{\text{real}}_{i,\,t},\ \mathbf{x}^{\text{sim}}_{j,\,t'}) \mid i \neq j,\ t \neq t' \big\}.
\end{split}
\end{equation*}
{Together, these negatives encourage the probe to detect fine-grained trajectory differences rather than rely on task-level cues, enabling detection of subtle motion inconsistencies at inference.}

\textbf{Probe training and inference.} We then use the collected positive and negative pairs to train the probe, where $\mathcal{P} = \mathcal{P}^{+} \cup \mathcal{P}^{-}$ denotes the training dataset. For a sampled pair $(\mathbf{x}^{\text{real}}_{i,\,t}, \mathbf{x}^{\text{sim}}_{j,\,t'}) \sim \mathcal{P}$, we uniformly sample 16 frames from each clip and encode them using a pre-trained video encoder to obtain embeddings $\mathbf{z}_{\text{real}}$ and $\mathbf{z}_{\text{sim}}$. The embeddings are concatenated and passed to the attention-based probe $g_\phi(\cdot)$. The probe consists of a single cross-attention layer with a learnable query token, followed by a classification head that outputs an alignment logit $\ell = g_\phi\!\left([\mathbf{z}_{\text{real}}, \mathbf{z}_{\text{sim}}]\right)$. With alignment probability $p = \sigma(\ell)$, we train $g_{\phi}$ with a binary cross-entropy loss:
\begin{align*}
\mathcal{L}(\phi;\mathcal{P})
= \mathbb{E}_{(\mathbf{x}^{\text{real}}_{i,\,t},\, \mathbf{x}^{\text{sim}}_{j,\,\smash[t]{t'}}) \sim \mathcal{P}}
\bigl[-y\log p - (1-y)\log(1-p)\bigr], \quad y := \mathbf{1}[i = j \wedge t = t'].
\end{align*} 
At inference, we first use a VLM to score the physical plausibility of robot arm and end-effector motion in each generated video based on depth and spatial orientation, and discard videos scoring below 3 out of 5 unless otherwise specified. For each synthetic data sample $(\mathbf{x}_{\text{gen}}, \mathbf{A}_{\text{IDM}})$ that passes this pre-filter, we form the video pair $(\mathbf{x}_{\text{gen}}, \hat{\mathbf{{x}}}_{\text{sim}})$ and retain the sample only if $g_{\phi}$ classifies the pair as positive. Unless otherwise stated, we use $p > 0.5$ as the decision boundary. {We provide further implementation details in~\Cref{appendix:attentive_probe_details}.}

\textbf{{Best-of-N sampling with generation-time critic.}} Beyond the post-hoc filtering strategy, our probe can also serve as a critic for the video generative model at generation time to improve the synthetic data quality. Specifically, we adopt a Best-of-N sampling strategy: (1) we sample $N$ candidate videos along with their IDM-predicted actions, (2) we apply the same VLM pre-filter used at probe inference to each candidate video, and (3) we select the video-action pair with the highest critic score among the remaining candidates. We use the alignment probability $p$ as the critic score, which measures the motion consistency between $\mathbf{x}_{\text{gen}}$ and  $\hat{\mathbf{x}}_{\text{sim}}$. A higher score indicates better alignment between $\mathbf{x}_{\text{gen}}$ and $\mathbf{A}_{\text{IDM}}$, making the selected pair more suitable for policy learning. This strategy is particularly appealing in fine-tuning settings, where data collection is typically constrained to a specific target task (see \Cref{tab:main_allex}). Importantly, Best-of-N sampling improves the synthetic data quality by applying action-verification at generation time, enabling more effective use of our action-verification framework in data-scarce settings.

\subsection{{{Scene} Diversification of Synthetic Data}}
\label{subsec:generate}

While our action-verified filtering discards low-quality samples, its effectiveness fundamentally depends on the diversity of the synthetic data. 
Existing synthetic data generation frameworks~\citep{jang2025dreamgen} focus on generating novel manipulation scenarios but inherit the limited scene diversity of the underlying real data. To diversify scenes, we decompose the manipulation scene into editing components such as (1) table appearance, (2) target object identity and appearance, (3) lighting, and (4) background, and diversify each component while preserving the rest. We apply this design at two complementary levels using off-the-shelf generative models: {image level (I2I editing)}, where we edit the initial frame to generate novel trajectories under varied scenes, and {video level (V2V transfer)}, where we transform previously generated trajectories to diversify the scene while preserving the underlying motion. In particular, we first query a VLM to extract the manipulation context from the initial frame and task instruction. For I2I editing, we then generate edited variants of the initial frame along these components, and feed them into the video generative model. For V2V transfer, we transform generated videos showing task success into new scenes and reuse their IDM-predicted action labels by preserving object shape and robot color via additional prompting. To enable more precise editing, both stages are additionally conditioned on Canny edge maps~\citep{ali2025world} that preserve the original scene structure. We provide implementation details in~\Cref{appendix:synthetic_data_generation} and prompt details in~\Cref{appendix:prompts_i2i,appendix:prompts_v2v}.

\section{Experiments}
\label{sec:experiments}

We design experiments to answer the following research questions:

\vspace{-0.3em}

\begin{itemize}[leftmargin=*, topsep=2pt, itemsep=1pt, parsep=0pt]
    \item How effective is the {scene} diversification pipeline in \Methodname{} (see~{\Cref{tab:main_gr1_tabletop,tab:main_dexmg}})?
    \item Does pre-training with \Methodname{} effectively improve VLA performance (see~\Cref{tab:main_gr1_tabletop,tab:main_dexmg})?
    \item Does co-finetuning with \Methodname{} effectively improve VLA performance (see~\Cref{tab:main_allex})?
    \item How effective is the action-verified filtering strategy in \Methodname{} (see~\Cref{tab:ablation_comparison,tab:ablation_component,tab:ablation_inconsistent_action,tab:ablation_encoder,tab:ablation_generalization,tab:ablation_threshold})?
\end{itemize}

\begin{figure}[tp!]
    \vspace{-0.1in}
    \centering
    \begin{minipage}[t]{0.48\textwidth}
        \vspace{0pt}
        \centering
        \includegraphics[width=0.78\linewidth]{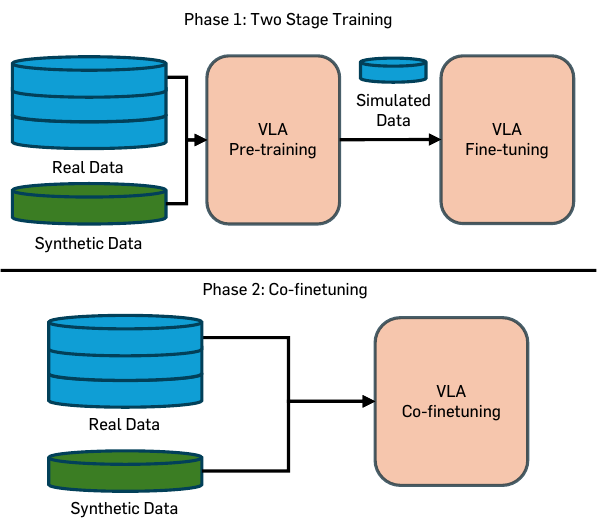}
        \caption{\textbf{An overview of experimental design for \Methodname.}
        We conduct two-phase experiments: (1) pre-training on real and synthetic data followed by fine-tuning on simulated data, and (2) co-finetuning on real and synthetic data.}
        \label{fig:experimental_design}
    \end{minipage}
    \hfill
    \begin{minipage}[t]{0.48\textwidth}
        \vspace{8pt}
        \centering
        \includegraphics[width=0.32\linewidth]{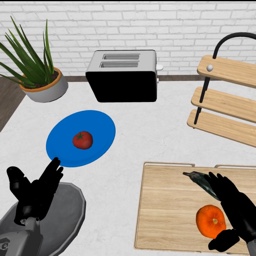}
        \includegraphics[width=0.32\linewidth]{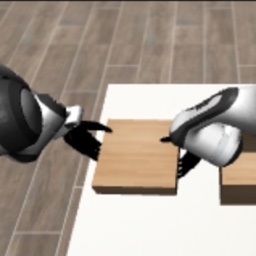}
        \includegraphics[width=0.32\linewidth]{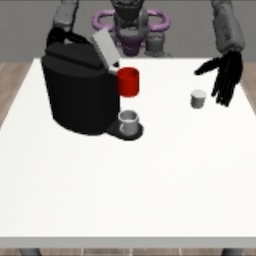} \\
        \includegraphics[width=0.32\linewidth]{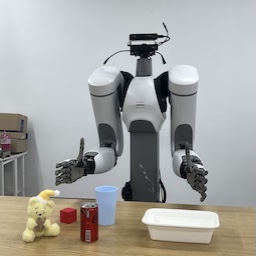}
        \includegraphics[width=0.32\linewidth]{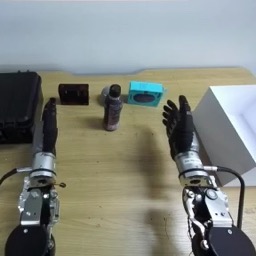}
        
        \caption{\textbf{Visualization of benchmarks.} Top: simulation benchmarks including (1) GR-1 Tabletop~\citep{nvidia2025gr00tn1openfoundation} and (2) DexMimicGen~\citep{jiang2025dexmimicgen} with bimanual Panda arms equipped with dexterous hands and (3) the GR-1 humanoid. Bottom: real-world benchmarks with the (4) ALLEX humanoid and (5) OpenArm humanoid.}
        \label{fig:benchmark_visualization}
    \end{minipage}
\vspace{-0.2in}
\end{figure}

We evaluate \Methodname and answer these questions under two training regimes: (1) two-stage training (pre-training followed by fine-tuning), and (2) co-finetuning on real and synthetic data. We provide a brief overview of our experimental design in~\Cref{fig:experimental_design} and visualize the evaluated benchmarks in~\Cref{fig:benchmark_visualization}.

\subsection{Pre-training Experiments}
\label{subsec:two_stage_train}

\textbf{Datasets.} For the pre-training stage, we use ActionNet~\citep{fourier2025actionnet}, a real robot demonstration dataset, as the default pre-training dataset. ActionNet consists of bimanual tabletop manipulation trajectories collected on the Fourier GR1-T1 humanoid, captured from  {an} ego-view camera. The GR1-T1 humanoid is equipped with {6-DoF} dexterous hands, resulting in a total of 44-dimensional joint-space states and actions. While the original dataset has 30K teleoperated episodes, we use a 3K subset for pre-training by removing episodes with duplicate task instructions. 

For the fine-tuning stage, we fine-tune and evaluate models on two {simulation dexterous manipulation benchmarks}: GR-1 Tabletop~\citep{nvidia2025gr00tn1openfoundation} and DexMimicGen~\citep{jiang2025dexmimicgen}. GR-1 Tabletop focuses on dexterous hand control and includes 24 tabletop tasks (18 rearrangement and 6 articulated) with a variety of objects and placements. DexMimicGen focuses on bimanual manipulation and {supports} cross-embodiment evaluation across 6 tasks (3 GR-1 humanoid and 3 bimanual Panda arms with dexterous hands tasks). For both benchmarks, we evaluate models over 50 trials per task across all experiments.

\textbf{Baselines.} We primarily use GR00T N1.5 \citep{nvidia2025gr00t} as the base policy and do not load the pre-trained weights for the action head. We design our baselines based on two factors: (1) the choice of pre-training data, and (2) the use of action-verified filtering. For pre-training data, we consider three settings: (1) real data only, (2) real data combined with synthetic data generated by the DreamGen \citep{jang2025dreamgen} pipeline without scene diversification, and (3) real data combined with scene-diversified synthetic data from \Methodname. We further apply our action-verified filtering to setting (3) to compare uncurated and curated versions. We provide full implementation details for policy training in~\Cref{appendix:implementation_pretrain}, and for synthetic data generation in~\Cref{appendix:synthetic_data_generation}.

\begin{table*}[t!]
\vspace{-0.1in}
\caption{\textbf{Performance comparison on GR-1 Tabletop \citep{nvidia2025gr00tn1openfoundation}}. We report the average success rate (\%) over 50 trials per task, fine-tuned with a varying number of demonstrations. DreamGen \citep{jang2025dreamgen} denotes synthetic data without our scene diversification pipeline. For a fair comparison, we pre-train all methods using synthetic data with the same dataset size of 10K samples before filtering.}
\centering
\fontsize{7pt}{8pt}\selectfont
\begin{tabular}{lcccccccc}
\toprule
\multirow{2.5}{*}{\makecell{Method}} &
\multirow{2.5}{*}{\makecell{Synth.}} &
\multirow{2.5}{*}{\makecell{Filtering}} &
\multicolumn{3}{c} {300 Demos} & \multicolumn{3}{c} {1000 Demos} \\ \cmidrule(lr){4-6} \cmidrule(lr){7-9} &&& 
Rearrangement & Articulated & \textbf{Avg.} & Rearrangement & Articulated & \textbf{Avg.} \\
\cmidrule(lr){1-3} \cmidrule(lr){4-9}
Real & \textcolor{SJRed}{\xmark} & \textcolor{SJRed}{\xmark} & 16.1 & 13.3 & 15.4  & 30.9 & 19.0 & 27.9 \\ 
w/ DreamGen~\citep{jang2025dreamgen} & \textcolor{SJViolet}{\cmark} & \textcolor{SJRed}{\xmark} & 21.1 & 14.7 & 19.5 & 30.7 & 36.3 & 32.1 \\
\rowcolor{gray!10}
\textbf{w/ \Methodname (Ours)} & \textcolor{SJViolet}{\cmark} & \textcolor{SJRed}{\xmark} & 23.2 & 21.0 & 22.7 & 31.0 & \textbf{43.0} & 34.0 \\
\rowcolor{gray!10}
\textbf{w/ \Methodname (Ours)} & \textcolor{SJViolet}{\cmark} & \textcolor{SJViolet}{\cmark} & \textbf{30.9} & \textbf{24.3} & \textbf{29.3} & \textbf{32.3} & 41.3 & \textbf{34.6} \\
\bottomrule
\end{tabular}
\label{tab:main_gr1_tabletop}
\vspace{-0.05in}
\end{table*}

\textbf{Experimental results.} \Cref{tab:main_gr1_tabletop,tab:main_dexmg} show that pre-training with our synthetic data framework consistently outperforms all other baselines. First, we observe that our scene diversification pipeline (\ie I2I editing and V2V transfer) substantially improves downstream task performance. Second, our action-verified filtering is effective for curating synthetic data and further enhances VLA performance, even with a smaller amount of  pre-training data. Finally, our synthetic data demonstrates cross-embodiment transfer, where a policy pre-trained on GR-1 humanoid synthetic data helps form a manipulation prior for bimanual Panda arms with dexterous hands.

\begin{table*}[t!]
\caption{\textbf{Performance comparison on DexMimicGen \citep{jiang2025dexmimicgen}}. We report the average success rate (\%) over 50 trials per task across 3 random seeds, fine-tuned with 100 demonstrations per task. DreamGen \citep{jang2025dreamgen} denotes synthetic data without our scene diversification pipeline. For a fair comparison, we pre-train all methods using synthetic data with the same dataset size of 10K samples before filtering.}
\centering
\fontsize{7pt}{8pt}\selectfont
\begin{tabular}{lccccc}
\toprule
Method & Synth. & Filtering &
GR-1 Humanoid & {Bimanual Panda Arms} & \textbf{Avg.} \\ 
\cmidrule(lr){1-3} \cmidrule(lr){4-6}
Real & \textcolor{SJRed}{\xmark} & \textcolor{SJRed}{\xmark}  
& 56.9 &  32.2 & 44.6  \\
w/ DreamGen~\citep{jang2025dreamgen} & \textcolor{SJViolet}{\cmark} & \textcolor{SJRed}{\xmark}  
& 57.1 & 35.6 & 46.4 \\
\rowcolor{gray!10}
\textbf{w/ \Methodname (Ours)} & \textcolor{SJViolet}{\cmark} & \textcolor{SJRed}{\xmark} 
&  59.3 & 39.3 & 49.3 \\
\rowcolor{gray!10}
\textbf{w/ \Methodname (Ours)} & \textcolor{SJViolet}{\cmark} & \textcolor{SJViolet}{\cmark} 
& \textbf{72.2} & \textbf{42.9} & \textbf{57.6} \\
\bottomrule
\end{tabular}
\label{tab:main_dexmg}
\end{table*}

\begin{table*}[t]
\caption{\textbf{Performance comparison on ALLEX and OpenArm.} We report the average success rate (\%) over 24 trials per task {on ALLEX and over 32 trials per task on OpenArm}. DreamGen \citep{jang2025dreamgen} denotes synthetic data without our scene diversification pipeline. * denotes Best-of-N sampling using our filtering strategy without scene diversification.}
\centering
\fontsize{7pt}{8pt}\selectfont
\setlength{\tabcolsep}{3.5pt}
\resizebox{\linewidth}{!}{%
\begin{tabular}{lcccccccccc}
\toprule
\multirow{3.5}{*}{Method} &
\multirow{3.5}{*}{Synth.} &
\multirow{3.5}{*}{Filtering}
& \multicolumn{4}{c}{ALLEX} & \multicolumn{3}{c}{{OpenArm}} \\
\cmidrule(lr){4-7}\cmidrule(lr){8-10}
&&& \makecell{PnP Can\\(In-distribution)} & \makecell{PnP Cup\\(Novel Object)} & \makecell{Pour Can\\(Novel Behavior)} & \textbf{Avg.}
& {\makecell{Lift Bottle\\(In-distribution)}} & {\makecell{Lift Box\\(Novel Object)}} & {\textbf{Avg.}} \\
\midrule
Real & \textcolor{SJRed}{\xmark} & \textcolor{SJRed}{\xmark} & 25.0 & 16.7 & 0.0 & 13.9
& {37.5} & {\phantom{0}6.3} & {21.9} \\
w/ DreamGen~\citep{jang2025dreamgen} & \textcolor{SJViolet}{\cmark} & \textcolor{SJRed}{\xmark} & 37.5 & 33.3 & 12.5 & 27.8
& {43.8} & {\phantom{0}9.4} & {26.6} \\
\rowcolor{gray!10}
\textbf{w/ RoboCurate* (Ours)} & \textcolor{SJViolet}{\cmark} & \textcolor{SJViolet}{\cmark} & \textbf{47.9} & \textbf{43.8} & \textbf{25.0} & \textbf{38.9}
& {\textbf{46.9}} & {\textbf{15.6}} & {\textbf{31.3}} \\
\bottomrule
\end{tabular}%
}
\label{tab:main_allex}
\end{table*}

\subsection{Co-finetuning Experiments}
\label{subsec:co_finetune}

\textbf{Datasets.}
We evaluate real-world tasks with two humanoid robots, ALLEX and OpenArm, to validate that synthetic data can improve real-world generalization without additional data collection for unseen tasks. The ALLEX humanoid is equipped with an ego-view camera and two 15-DoF hands, yielding 48-dimensional joint-space states and actions. The OpenArm humanoid similarly has an ego-view camera and two 6-DoF Inspire hands, with 28-dimensional joint-space states and actions. Specifically, we conduct experiments on 3 tabletop manipulation tasks on ALLEX and 2 tabletop lifting tasks on OpenArm (see~\Cref{tab:main_allex}). For each robot, we manually collect teleoperated demonstrations only for the in-distribution (ID) task (48 for ALLEX and 44 for OpenArm), while the out-of-distribution (OOD) tasks involve a novel object or a novel behavior and use no real data.

\textbf{Baselines.}
{We fine-tune pre-trained GR00T N1.5~\citep{nvidia2025gr00t} and RLDX-1~\citep{rldx2026} as the base policies for ALLEX and OpenArm humanoids, respectively.} We design baselines to evaluate our action-verification as a generation-time critic. {For co-finetuning data, we consider three settings:} (1) real data only, {(2) real data combined with synthetic data generated using the DreamGen \citep{jang2025dreamgen}} pipeline, and (3) real data combined with synthetic data selected by Best-of-N sampling using our filtering strategy as a critic. For Best-of-N sampling in (3), we generate 4 candidates with different Gaussian noise seeds; in contrast, (2) uses a single sample with a fixed seed. In (2) and (3), we generate 48 episodes for the ID task and 50 for each OOD task {on ALLEX, and 44 episodes for the ID task and 60 for the OOD task on OpenArm}.  Since {the co-finetuning phase} requires task-specific data, we generate synthetic data using a fixed task instruction per task, following the pipeline in~\Cref{appendix:synthetic_data_generation}.  We provide full implementation details for policy training in~\Cref{appendix:implementation_cofinetune}.

\textbf{Experimental results.} 
~\Cref{tab:main_allex} shows that \Methodname consistently outperforms the baselines on both humanoid benchmarks in the co-finetuning setup. On the ALLEX humanoid, our synthetic data enables the policy to perform a novel behavior that the real-data-only baseline cannot achieve (0.0\% $\rightarrow$ 25.0\% {on} pour can). It also doubles the success rate over synthetic data without Best-of-N sampling (12.5\% $\rightarrow$ 25.0\%).
{These trends also hold on OpenArm, where \Methodname yields a +42.9\% average relative gain over the real-data-only baseline, with the largest relative gain on the novel object task (6.3\% $\rightarrow$ 15.6\% {on} lift box).} Together, these results indicate that our filtering strategy selects better-aligned synthetic samples from the generated candidates, improving downstream task performance.

\begin{table*}[t!]
\vspace{-0.1in}
\centering
\begin{minipage}[t]{0.45\textwidth}
\caption{\textbf{Comparison with other filtering strategies} on GR-1 Tabletop \citep{nvidia2025gr00tn1openfoundation}. We compare our action-verified filtering strategy with DreamGenBench \citep{jang2025dreamgen} and VideoCon-Physics \citep{bansal2024videophy}, which assess the physical plausibility of generated videos with VLMs.}
\label{tab:ablation_comparison}
\centering
\vspace{0.31in}
\fontsize{8pt}{10pt}\selectfont
\resizebox{\linewidth}{!}{%
\begin{tabular}{lcc}
\toprule 
Method & Filtering & \textbf{Avg.} \\
\midrule
Real w/ Synthetic & \textcolor{SJRed}{\xmark} & 22.7 \\
+ DreamGenBench~\citep{jang2025dreamgen} & \textcolor{SJViolet}{\cmark} & 22.8 \\
+ VideoCon-Physics~\citep{bansal2024videophy} & \textcolor{SJViolet}{\cmark} & 23.0  \\
\rowcolor{gray!10}
+ \textbf{\Methodname (Ours)} & \textcolor{SJViolet}{\cmark} & \textbf{29.3} \\
\bottomrule
\end{tabular}%
}
\end{minipage}%
\hfill
\begin{minipage}[t]{0.51\textwidth}
\caption{\textbf{Component-wise ablation of action-verified filtering} on GR-1 Tabletop~\citep{nvidia2025gr00tn1openfoundation}. Our filtering strategy consists of a VLM pre-filtering stage followed by video embedding comparison. The top-$K$ filter variant ranks samples by cosine similarity between their video embeddings and retains $K$ samples, matching the number kept by the attentive probe used in \Methodname.}
\label{tab:ablation_component}
\centering
\fontsize{8pt}{10pt}\selectfont
\resizebox{\linewidth}{!}{%
\begin{tabular}{lcc}
\toprule
Method & Filtering & \textbf{Avg.} \\
\midrule
Real w/ Synthetic & \textcolor{SJRed}{\xmark} 
& 22.7 \\
+ VLM pre-filter & \textcolor{SJViolet}{\cmark} & 24.2 \\
\quad + Top-$K$ filter w/ cosine sim. & \textcolor{SJViolet}{\cmark} & 26.0 \\
\rowcolor{gray!10}
\quad \textbf{+ Attentive probe (Ours)} & \textcolor{SJViolet}{\cmark} & \textbf{29.3} \\
\bottomrule
\end{tabular}%
}
\end{minipage}
\vspace{-0.2in}
\end{table*}

\subsection{Ablation Studies and Analyses}
\label{subsec:ablation_studies}

We investigate the effectiveness of the proposed components of \Methodname under the two-stage training setup shown in~\Cref{fig:experimental_design}. We use ActionNet \citep{fourier2025actionnet} as the default pre-training dataset and GR00T N1.5 \citep{nvidia2025gr00t} as the base policy, without loading pre-trained weights for the action head. Unless otherwise stated, we use our scene-diversified synthetic data for pre-training, fix its size to 10K samples before filtering, and fine-tune with 300 demonstrations per task on GR-1 Tabletop~\citep{nvidia2025gr00tn1openfoundation}. Additional implementation details are provided in \Cref{appendix:ablation_details}.

\begin{wraptable}{r}{0.40\textwidth}
\vspace{-0.18in}
\caption{\textbf{Analysis of inconsistent action detection.} We report the percentage of samples correctly discarded by each filtering strategy.} 
\centering
\fontsize{8pt}{10pt}\selectfont
\begin{tabular}{lc}
\toprule
Method & Discard Rate (\%) \\
\midrule
DreamGenBench~\citep{jang2025dreamgen} & \phantom{0}3.7 \\
\rowcolor{gray!10}
\textbf{\Methodname (Ours)} & \textbf{70.4} \\
\bottomrule
\end{tabular}
\label{tab:ablation_inconsistent_action}
\end{wraptable}

\textbf{Comparison with other filtering strategies.}
To verify the effectiveness of our filtering strategy, we compare our filter with other methods that assess the physical plausibility of generated videos.~\Cref{tab:ablation_comparison} {shows} that our action-verified filtering clearly outperforms video plausibility filters. In particular, DreamGenBench \citep{jang2025dreamgen} queries a {VLM} with a fixed prompt to judge whether a generated video violates basic physics. Similarly, VideoCon-Physics \citep{bansal2024videophy} queries VideoCon, a 7B {VLM}, to predict whether a video follows physical laws. The results demonstrate that VLM-based video plausibility assessments alone are insufficient for curating beneficial synthetic data, and that verifying consistency between generated videos and their action labels helps assess action quality, which is crucial for policy learning.
To better understand this gap, we analyze a specific failure mode where the generated video is accurate but the IDM-predicted action is inconsistent (see~\Cref{fig:av_ia_examples}). As shown in \Cref{tab:ablation_inconsistent_action}, we find that our filter effectively removes inconsistent action samples, while the VLM-based filter almost entirely accepts these failure cases since it does not inspect actions, supporting the results of~\Cref{tab:ablation_comparison}. We note that we manually label 400 synthetic samples and identify 27 such failure cases for evaluation.

\textbf{Effect of action-verified filter design.} 
We further investigate the components of our action-verified filtering strategy.~\Cref{tab:ablation_component} decomposes it into two stages: (1) VLM pre-filtering and (2) video embedding comparison. For the second stage, we compare a Top-$K$ filter based on cosine similarity between embeddings of the generated video and simulator replay with our attentive probe. We observe that our attentive probe outperforms the Top-$K$ filter (26.0\% to 29.3\%), suggesting that training on our constructed positive and negative pairs (see~\Cref{fig:probe_train_data}) helps it detect motion inconsistencies more effectively than raw embedding similarity. Additionally, the VLM pre-filter alone raises the average success rate by 1.5 percentage points (22.7\% to 24.2\%), while adding the attentive probe yields a further gain of 5.1 points (24.2\% to 29.3\%), highlighting the value of action-verification for policy learning with synthetic data.

\begin{wraptable}{r}{0.48\textwidth}
\vspace{-1.30em}
\caption{{\textbf{Effect of video encoder in attentive probe} on GR-1 Tabletop~\citep{nvidia2025gr00tn1openfoundation}. We report the average success rate (\%) over 50 trials per task.}}
\label{tab:ablation_encoder}
\centering
\fontsize{8pt}{10pt}\selectfont
\resizebox{\linewidth}{!}{%
\begin{tabular}{llcc}
\toprule
Method & Video Encoder & Filtering & \textbf{Avg.} \\
\midrule
Real w/ Synthetic & - & \textcolor{SJRed}{\xmark} & 22.7 \\
\rowcolor{gray!10}
+ \Methodname (Ours) & VideoMAEv2~\citep{wang2023videomaev2} & \textcolor{SJViolet}{\cmark} & 25.4 \\
\rowcolor{gray!10}
\textbf{+ \Methodname (Ours)} & V-JEPA 2~\citep{assran2025v} & \textcolor{SJViolet}{\cmark} & \textbf{29.3} \\
\bottomrule
\end{tabular}%
}
\vspace{-1.0em}
\end{wraptable}

\textbf{Effect of video encoder in attentive probe.}
To study the effect of the pre-trained video encoder in our probe, we replace V-JEPA 2~\citep{assran2025v} (300M) with the smaller VideoMAEv2~\citep{wang2023videomaev2} (83M), changing only encoder-specific probe training settings (\eg input resolution). As shown in~\Cref{tab:ablation_encoder}, our filtering strategy outperforms the unfiltered baseline with both video encoders, and the larger V-JEPA 2 encoder yields a greater gain. This suggests that our filtering strategy is robust across encoder choices and may benefit from stronger video encoders.

\begin{table*}[t!]
\vspace{-0.1in}
\centering
\begin{minipage}[t]{0.48\textwidth}
\caption{\textbf{Effect of action-verified filtering on other synthetic data} on GR-1 Tabletop \citep{nvidia2025gr00tn1openfoundation}. We report the average success rate (\%) over 50 trials per task, and use 0.3K real and 1K synthetic samples for pre-training. We note that we use synthetic data without scene diversification for both methods.}
\label{tab:ablation_generalization}
\centering
\fontsize{8pt}{10pt}\selectfont
\resizebox{\linewidth}{!}{%
\begin{tabular}{lccc}
\toprule
Method & Scene Div. & Filtering & \textbf{Avg.} \\
\midrule
Real w/ Synthetic & \textcolor{SJRed}{\xmark} & \textcolor{SJRed}{\xmark} & 15.8 \\
\rowcolor{gray!10}
\textbf{+ RoboCurate (Ours)} & \textcolor{SJRed}{\xmark} & \textcolor{SJViolet}{\cmark} & \textbf{20.0} \\
\bottomrule
\end{tabular}%
}
\end{minipage}%
\hfill
\begin{minipage}[t]{0.48\textwidth}
\caption{{\textbf{Effect of attentive probe threshold}} on GR-1 Tabletop~\citep{nvidia2025gr00tn1openfoundation}. We report the average success rate (\%) over 50 trials per task. The gray row indicates our default setup.}
\label{tab:ablation_threshold}
\centering
\vspace{0.18in}
\fontsize{8pt}{10pt}\selectfont
\resizebox{\linewidth}{!}{%
\begin{tabular}{lccc}
\toprule
Method & Filtering & Threshold & \textbf{Avg.} \\
\midrule
Real w/ Synthetic & \textcolor{SJRed}{\xmark} & -- & 22.7 \\
+\textbf{ \Methodname (Ours)} & \textcolor{SJViolet}{\cmark} & 0.10 & 25.2 \\
+ \textbf{\Methodname (Ours)} & \textcolor{SJViolet}{\cmark} & 0.25 & 26.6 \\
\rowcolor{gray!10}
+ \textbf{\Methodname (Ours)} & \textcolor{SJViolet}{\cmark} & 0.50 & \textbf{29.3} \\
\bottomrule
\end{tabular}%
}
\end{minipage}
\vspace{-0.1in}
\end{table*}

\textbf{Effect of threshold on attentive probe.}  
To investigate sensitivity to the amount of synthetic data retained, we vary the attentive probe threshold. As shown in~\Cref{tab:ablation_threshold}, our method outperforms the baseline using unfiltered synthetic data at thresholds from 0.10 to 0.50, including the default value of 0.50. This suggests that the effectiveness of \Methodname is robust to the choice of probe threshold. We provide further analysis in~\Cref{appendix:attentive_probe_analysis}.

\textbf{Generalization of action-verified filtering.} 
{To verify the generalizability of our filtering strategy, we apply our filter to synthetic data generated with the DreamGen~\citep{jang2025dreamgen} pipeline without our scene diversification. \Cref{tab:ablation_generalization} shows that our action-verified filtering improves policy performance, demonstrating that our filter enhances synthetic data quality regardless of whether scene diversification is adopted. We note that we use 0.3K real data samples and 1K synthetic data samples for pre-training.}

\section{Related Work}
\label{sec:related_work}

\textbf{Synthetic data generation for robot policies.} Collecting large-scale real-world robot data is costly and labor-intensive, which has led many works to rely on simulated datasets as a primary alternative \citep{mandlekar2023mimicgen, nasiriany2024robocasa}. While simulation enables scalable data collection, it suffers from several limitations, including the sim-to-real gap~\citep{zhao2020sim}, inaccurate physical modeling, and complex tool interactions \citep{zhu2022challenges}. As an alternative, recent works \citep{yuan2025roboengine, dong2025emma, liu2025robotransfer} generate synthetic robot data using generative models, aiming to bypass simulation constraints and improve scalability \citep{mandi2022cacti, alhaija2025cosmos}. For example, \citet{jang2025dreamgen} first sample plausible task instructions from an initial image, then generate synthetic robot videos using image-to-video (I2V) diffusion models, and infer pseudo-action labels using IDMs. However, these approaches can produce low-quality synthetic data: generated videos may be physically implausible, or IDM-predicted actions may fail to match the motion shown even in plausible videos, resulting in inaccurate action labels that harm policy learning. To the best of our knowledge, \Methodname is the first to explicitly assess the consistency of action labels with generated videos when curating synthetic robot data. We assess this consistency through simulator replay and demonstrate that such curation improves downstream policy performance.

\section{Conclusion}
\label{sec:conclusion}

In this work, we present \Methodname, a novel synthetic data curation framework for robot learning via action-verification. Our approach filters low-quality samples through simulator replay, transforming the challenging action-verification problem into a video motion-consistency problem. We additionally propose a scene diversification pipeline leveraging off-the-shelf generative models and use a VLM to pre-filter physically implausible videos, further supporting our action-verified filtering. This framework mitigates video-action misalignment in synthetic data from recent generative models and consistently improves VLA performance across a wide range of simulation and real-world benchmarks, including two real humanoids with dexterous hands. We believe \Methodname encourages principled quality evaluation for synthetic robot data and enables more effective policy learning in the data-scarce robotics domain.

\bibliography{example_paper}

\newpage
\appendix
\crefalias{section}{appendix}
\crefalias{subsection}{subappendix}
\crefalias{subsubsection}{subsubappendix}
\section{Discussion}
\label{appendix:discussion}

\subsection{Future works and limitations}
\label{appendix:future_works_limitations}
As future work, we believe it would be interesting to extend our action-verification framework beyond physics simulators by leveraging world models to generate rollout videos, enabling motion-consistency verification through video-to-video comparison and potentially mitigating the current sim-to-real domain gap. Furthermore, while our current scene diversification pipeline relies on pre-defined editing components, future advances in controllable video generation may enable more fine-grained and compositional scene editing. We also believe that analyzing which scene factors should be diversified at a system level could further improve the effectiveness of synthetic data generation for robot learning.

A potential limitation of \Methodname{} is that training the attentive probe requires real-world demonstrations to construct positive and negative pairs, which may be difficult to obtain in extremely low-data regimes. In addition, deploying our filtering framework as a Best-of-$N$ critic improves synthetic data quality but introduces additional generation cost, leading to a trade-off between filtering quality and inference efficiency in practical settings.

\subsection{Broader Impacts}
\label{appendix:broader_impact}
This paper presents \Methodname{}, a framework for improving the quality of synthetic robot data through action-verification. We expect that our approach may positively impact scalable robot learning by reducing the cost and effort of real-world robot data collection, potentially improving accessibility for resource-limited research groups. In addition, more reliable synthetic data generation could support safer and more robust training of VLAs for real-world manipulation tasks. 

One possible negative impact is that policies trained heavily on synthetic data may still inherit distribution gaps or unexpected failure modes during deployment, despite improved filtering quality. We therefore believe additional real-world evaluation and safety verification remain important before deploying such systems in safety-critical environments.

\clearpage
\section{Implementation Details}
\label{appendix:implementation_details}

\subsection{Pre-training Experiments} 
\label{appendix:implementation_pretrain}

For the pre-training stage, we adopt the Warmup-Stable-Decay (WSD; \citep{hu2024minicpm}) learning rate scheduler for policy training. Specifically, we initialize the action head from scratch and pre-train the policy for 60K gradient steps. The first 50K steps consist of a brief warmup followed by a constant learning rate, and the final 10K steps apply a linear decay. We hypothesize that the quality of synthetic data becomes more critical once the model has observed a sufficient amount of data. Therefore, we use all synthetic data for the first 50K steps and switch to only the synthetic data curated by \Methodname for the final 10K steps. We also use separate embodiment tags for real data (ActionNet \citep{fourier2025actionnet}) and synthetic data, despite their shared GR-1 embodiment, because the statistics of IDM-predicted actions differ from those of ActionNet. During training, we sample real and synthetic data in a 1:1 ratio by default. We otherwise follow the original training configuration of GR00T N1.5 \citep{nvidia2025gr00t}, except that global batch size is set to 512 and training is run for 60K gradient steps.

For the fine-tuning stage, we fine-tune GR00T N1.5 from the pre-trained checkpoint corresponding to each experimental setup (see Section~\ref{subsec:two_stage_train}). We note that the fine-tuning dataset is identical across runs for a given downstream task. We follow the original GR00T N1.5 fine-tuning configuration, except that we set the global batch size to 256 and tune the number of gradient steps per downstream task dataset (up to 60K) to optimize performance. We train separate models for different embodiments and observation viewpoints (\eg GR-1 humanoid in ego view, GR-1 humanoid in front view, and bimanual Panda arms with dexterous hands) on DexMimicGen \citep{jiang2025dexmimicgen}.

\subsection{Co-finetuning Experiments}
\label{appendix:implementation_cofinetune}

{\textbf{ALLEX humanoid experiments.} We follow the training setup of GR00T N1.5 \citep{nvidia2025gr00t} and fine-tune a pre-trained GR00T N1.5 model for 10K gradient steps. During co-finetuning, we mix real and synthetic data and maintain a 1:1:1 sampling ratio across pick-and-place can (ID), pick-and-place cup (OOD), and pour can (OOD) tasks. We otherwise follow the original GR00T N1.5 fine-tuning configurations, except global batch size is set to 128 and the number of gradient steps is set to 10K. During evaluation, we score each trial as 1 for success, 0.5 for partial success, and 0 for failure, and report the average success rate over 24 trials per task.

{\textbf{OpenArm humanoid experiments.} We fine-tune a pre-trained RLDX-1 checkpoint (RLDX-1-PT-IMG)~\citep{rldx2026} for 10K gradient steps with a global batch size of 64. During co-finetuning, we sample the real ID data, synthetic ID data and synthetic OOD data at a ratio of 1:1:1. We use separate normalization statistics for real and synthetic data for the action head, drawn from teleoperated demonstrations and the IDM training data, respectively. We evaluate each task over 32 trials, with 8 trials in each of 4 spatial regions of the workspace. Each trial is scored as 1 for success and 0 for failure.

\subsection{Attentive Probe Details}
\label{appendix:attentive_probe_details}

\textbf{Simulator replay details.} For GR-1 humanoid data, we replay recorded actions in a RoboCasa~\citep{nasiriany2024robocasa} simulation environment built on top of robosuite~\citep{zhu2020robosuite}. Specifically, we use the \texttt{GR1ArmsAndWaistFourierHands} robot embodiment within the \texttt{Tabletop} base environment with no objects, controlled at 20\,Hz. The scene uses a fixed layout and style (\texttt{layout\_id=0}, \texttt{style\_id=0}) to ensure visual consistency across replays; all other environment parameters follow their default values.

For ALLEX humanoid data, we replay recorded actions in an NVIDIA Isaac Sim~\citep{NVIDIA_Isaac_Sim} simulation environment. Specifically, we use the ALLEX robot embodiment (48 actuated DoFs and 60 total joints) loaded as a USD articulation, controlled at 20\,Hz. The scene contains only the robot against Isaac Sim's default empty stage (a uniform blue background with a grid floor) and no other objects, ensuring visual consistency across replays. Videos are rendered from an egocentric ZED MINI left-eye virtual camera rigidly attached to the neck link (\texttt{Neck\_Yaw\_link}), using pinhole intrinsics matched to the physical ZED MINI.

For OpenArm humanoid data, we replay recorded actions in the NVIDIA Isaac Lab~\citep{mittal2025isaac} simulation environment with the OpenArm embodiment (28-DoFs), controlled at 20\,Hz. Each video is rendered from an egocentric camera viewpoint and shows only the robot,  with no other objects.}

\textbf{Attentive probe training details.}
We train an attentive probe on top of a frozen V-JEPA 2 Large (300M) backbone \citep{assran2025v}, following the attentive probe architecture from V-JEPA 2. For each video in a pair, we uniformly sample 16 frames from a chosen frame interval. The two clips are independently encoded by the frozen V-JEPA 2 model, and the resulting token sequences are concatenated before being passed to the attentive probe. Training pairs consist of 80\% positives and 20\% negatives. Within the negatives, temporal shift and cross-episode negatives each constitute 45\%, with the remaining 10\% combining both strategies. For positives and cross-episode negatives, each clip's interval spans its full duration. For temporal shift negatives, the two clips share the same endpoint but begin at different start indices separated by at least 5 frames. Training hyperparameters are detailed in \Cref{table:attentive_probe_min_hparams}.

\vspace{12pt}
\begin{table*}[h!]
  \centering
  \caption{Attentive probe training hyperparameters.}
  \label{table:attentive_probe_min_hparams}
  \small
  \begin{tabular}{l c}
    \toprule
    \textbf{Hyperparameter} & \textbf{Value} \\
    \midrule
    Cross-attn heads ($n_{\text{heads}}$) & {8} \\
    Cross-attn layers ($L$)               & {1} \\
    Clip length (frames per video)        & {16} \\
    Input resolution                      & {256$\times$256} \\
    Optimizer                             & AdamW \\
    Learning rate                         & {1e-4} \\
    Weight decay                          & {0.05} \\
    Batch size (pairs)                    & {72} \\
    Epochs                                & {50} \\
    Positive / Negative ratio             & {80\% / 20\%} \\
    \bottomrule
  \end{tabular}
\end{table*}
\vspace{0.2in}

\textbf{VLM pre-filtering.}
To control the initial quality of generated videos, we check whether robot hand trajectories are plausible, with a correct sense of depth and spatial orientation. We prompt a VLM to score each video on a Likert scale from 1 to 5 and retain videos scoring at least 3 by default, raising the threshold to 4 for OpenArm experiments. We use Gemini 2.5 Flash \citep{comanici2025gemini} as a base model, feeding 8 video frames as input. The specific prompts are provided in \Cref{appendix:physical_plausibility}. We note that this pre-filter precedes action-verified filtering and the unfiltered baselines do not use it. We isolate its effect in~\Cref{tab:ablation_component}.}

\subsection{Computing Infrastructure}
\label{appendix:computing_infrastructure}
All experiments are conducted on NVIDIA A100 and H100 GPUs. Pre-training (\Cref{subsec:two_stage_train}) runs for 60K gradient steps with a global batch size of 512 on 4 NVIDIA A100 GPUs, requiring approximately 126 GPU hours in total. Co-finetuning (\Cref{subsec:co_finetune}) trains a single policy for 10K gradient steps in each setting. For ALLEX, we use a global batch size of 128 on 2 NVIDIA H100 GPUs, requiring approximately 4 GPU hours; for OpenArm, we use a global batch size of 64 on 2 NVIDIA A100 GPUs, requiring approximately 4 GPU hours. Fine-tuning in two-stage training~(\Cref{subsec:two_stage_train})  and ablation studies (\Cref{subsec:ablation_studies}) span multiple downstream tasks and configurations on 2 NVIDIA A100 GPUs, requiring approximately 52 GPU hours.

\clearpage
\section{Synthetic Robot Data Generation}
\label{appendix:synthetic_data_generation}

\textbf{Video generation.} We use Cosmos-Predict2-14B \citep{nvidia2025cosmospredict2} as the base video generation model except OpenArm, for which we use Cosmos-Predict2.5-2B~\citep{ali2025world}. We fine-tune Cosmos-Predict2-14B pre-trained checkpoint on ActionNet \citep{fourier2025actionnet} and GR00T-GR1-100\footnote{\url{https://huggingface.co/datasets/nvidia/PhysicalAI-Robotics-GR00T-GR1}} \citep{jang2025dreamgen} for the pre-training experiments, and on the manually collected ALLEX dataset for the co-finetuning. For OpenArm co-finetuning, we fine-tune Cosmos-Predict2.5-2B~\citep{ali2025world} pre-trained checkpoint on the manually collected OpenArm dataset. Unless otherwise stated, we generate 10K tabletop manipulation videos before filtering in the pre-training setup. We generate each video by conditioning the model on an initial image sampled from a fine-tuning dataset and a plausible task instruction. To produce these instructions, we prompt Gemini 3 Pro \citep{gemini3} with a carefully designed system prompt for each initial image, in the pre-training setup. We provide further details in  \Cref{appendix:i2v_task_instruction}. 

In addition, we use FLUX.2-dev~\citep{flux-2-2025} for image-to-image (I2I) editing and Cosmos-Transfer2.5-2B~\citep{ali2025world} for video-to-video (V2V) transfer as visual augmentation (see Section~\ref{subsec:generate}), except in the co-finetuning experiments. In our setup, I2I editing operates on initial images from the same dataset used for video generation, while V2V transfer operates on a subset of the previously generated videos. We generate approximately 10K augmented tabletop manipulation videos in total, with an I2I:V2V ratio of 2:1 for generation cost efficiency.

\textbf{Language re-captioning.} Inspired by \citet{nvidia2025gr00tn1openfoundation}, we apply instruction-following post-processing to our synthetic data in two stages. First, we prompt the VLM to check whether each video correctly follows its initial instruction, flagging those that do not. Second, we re-caption the flagged videos by prompting the VLM to generate a new instruction that describes the action depicted in the video. Because instruction following directly determines task success, we apply this post-processing to all synthetic datasets by default, yielding exactly 10K synthetic videos after this stage (except in the co-finetuning experiments). We use Gemini 3 Pro \citep{gemini3} as the VLM in both stages, feeding 8 video frames as input. The specific prompts are provided in \Cref{appendix:post_processing}.

\textbf{Pseudo-action labeling.} We use a diffusion transformer with a SigLIP-2 vision encoder \citep{tschannen2025siglip} as our base IDM, following \citet{jang2025dreamgen}. For the pre-training setup, we use an open-source pre-trained checkpoint\footnote{\url{https://huggingface.co/seonghyeonye/IDM_gr1}} for the GR-1 embodiment. For the co-finetuning setup, we train the IDM from scratch on {each robot} using manually collected data, with a global batch size of 64 for 20K gradient steps {on ALLEX humanoid and a global batch size of 128 for 30K gradient steps on OpenArm humanoid}.

\clearpage
\section{Examples of Synthetic Robot Data}
\label{appendix:synthetic_data_example}

\vspace{10pt}
\begin{figure}[!htbp]
    \centering
    \includegraphics[width=0.24\textwidth]{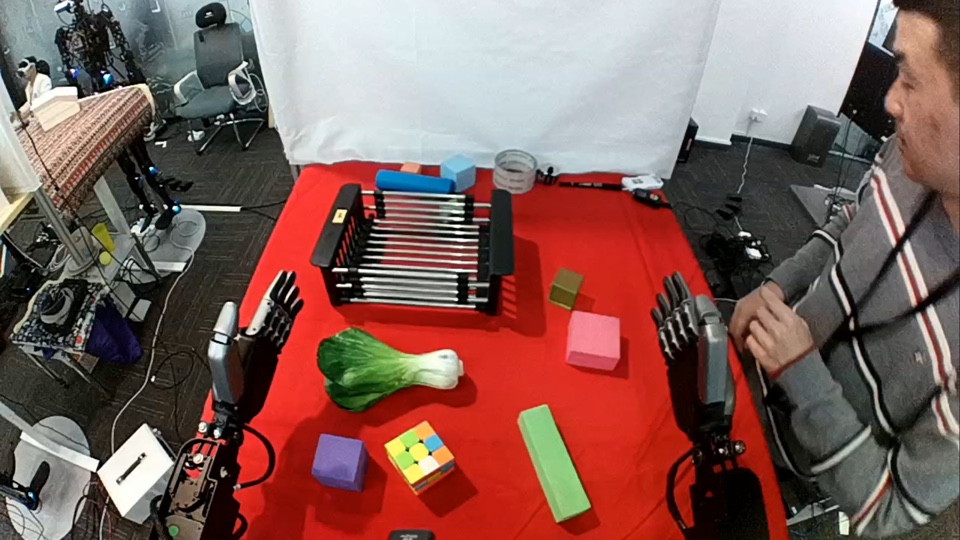}
    \includegraphics[width=0.24\textwidth]{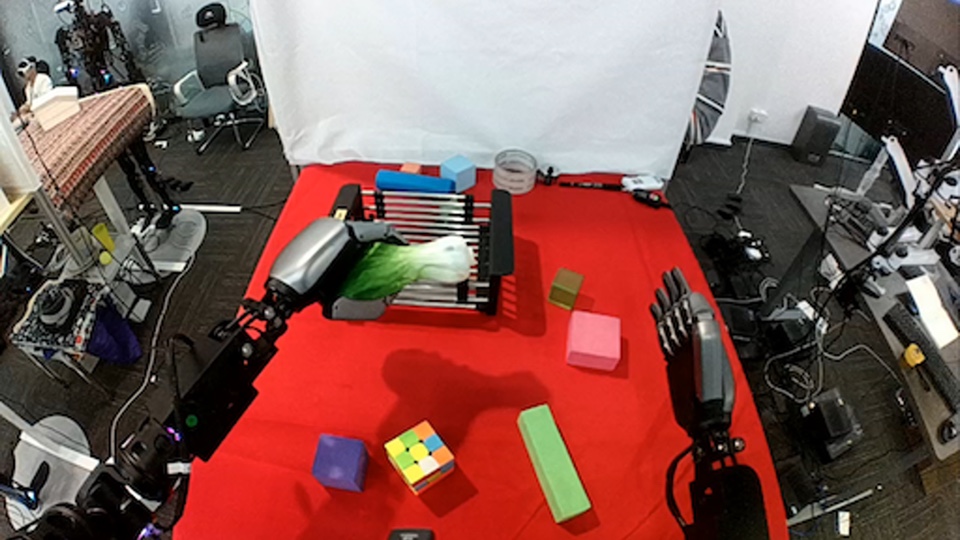}
    \includegraphics[width=0.24\textwidth]{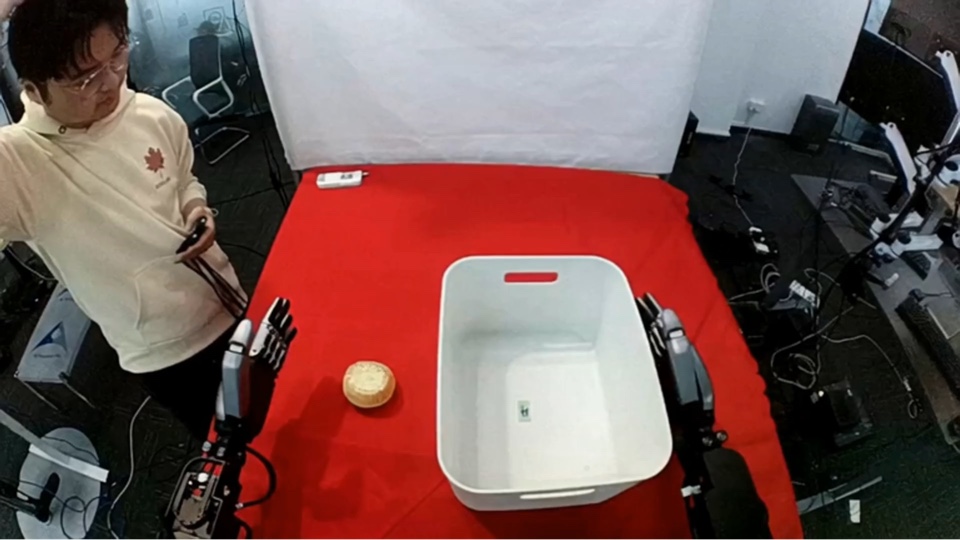}
    \includegraphics[width=0.24\textwidth]{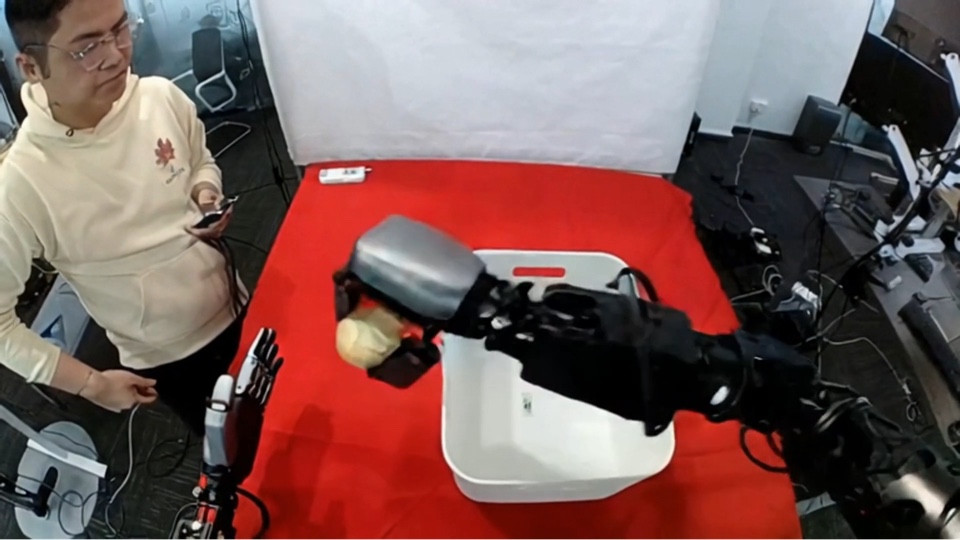}
    \\
    \includegraphics[width=0.24\textwidth]{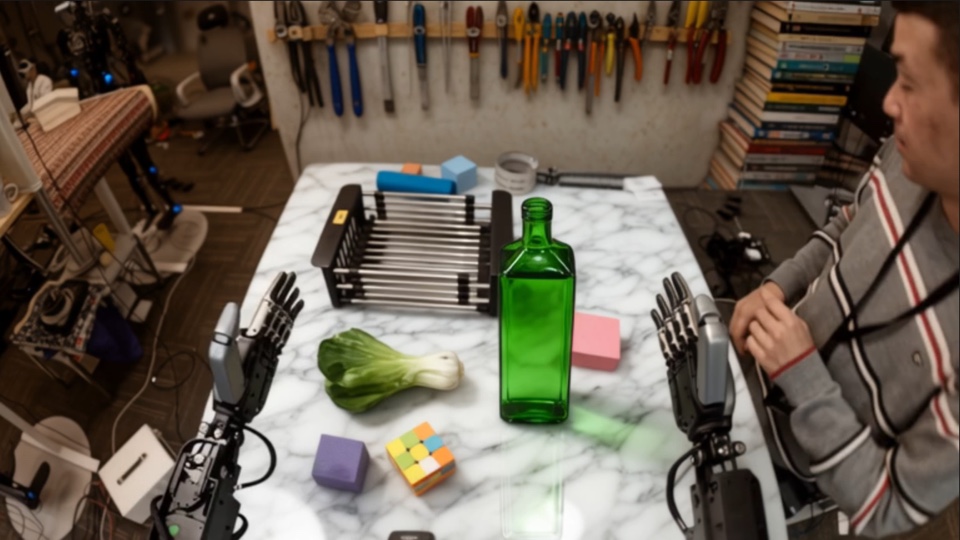}
    \includegraphics[width=0.24\textwidth]{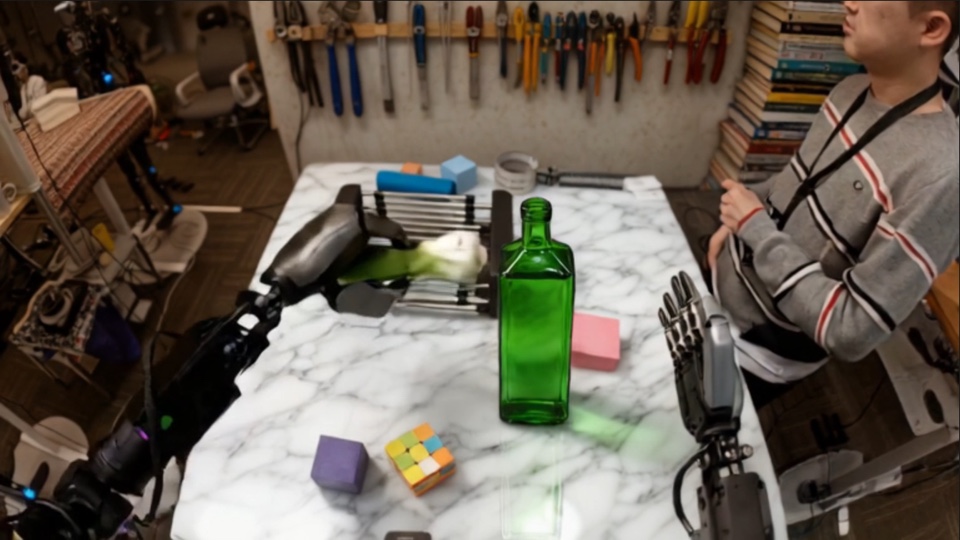}
    \includegraphics[width=0.24\textwidth]{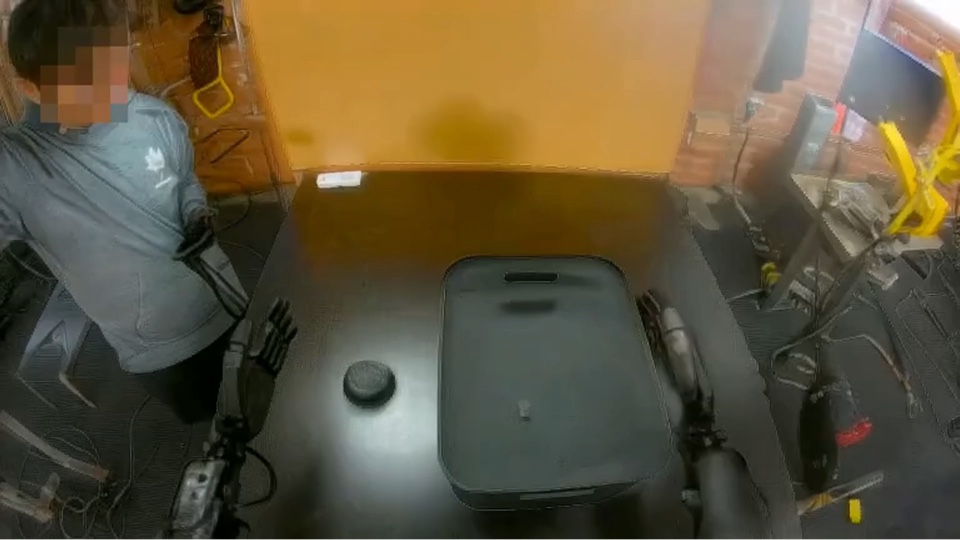}
    \includegraphics[width=0.24\textwidth]{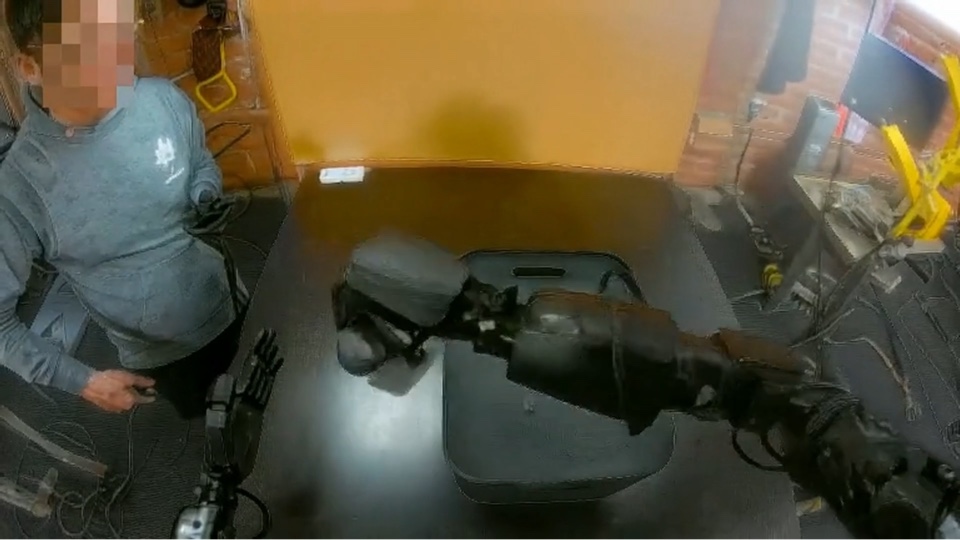}
    \caption{
    \textbf{Examples of synthetic data.} (Top): original videos, (Bottom): visually augmented synthetic data. The bottom-left pair shows a video whose initial frame was edited by the I2I model, while the bottom-right pair shows a video processed with V2V transfer.
    }
    \label{fig:synthetic_data_example}
\end{figure}

\vspace{20pt}

\begin{figure}[!htbp]
    \centering
    \includegraphics[width=0.24\textwidth]{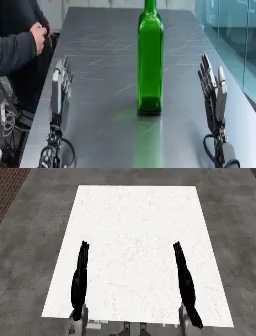}
    \includegraphics[width=0.24\textwidth]{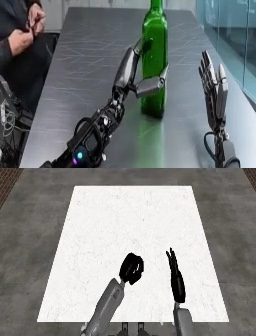}
    \includegraphics[width=0.24\textwidth]{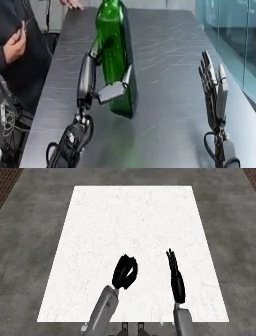}
    \includegraphics[width=0.24\textwidth]{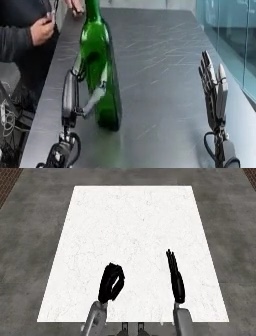}
    \\
    \includegraphics[width=0.24\textwidth]{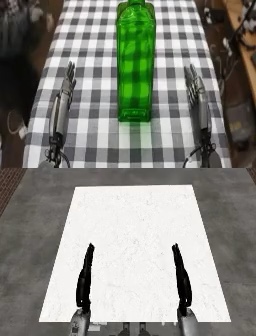}
    \includegraphics[width=0.24\textwidth]{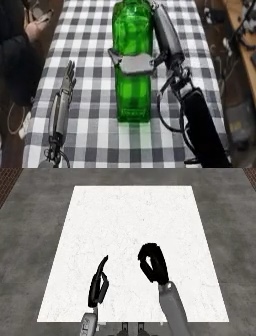}
    \includegraphics[width=0.24\textwidth]{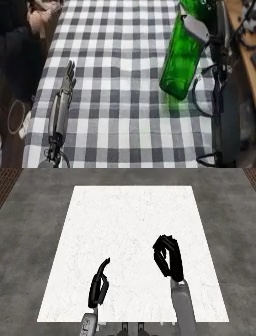}
    \includegraphics[width=0.24\textwidth]{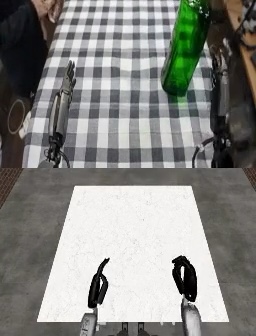}
    \caption{
    \textbf{Examples of accurate video with inconsistent action.} Each pair shows a generated video alongside the corresponding simulator replay produced from its IDM-predicted action (left to right). Although the generated videos are physically plausible, the IDM-predicted actions are inconsistent, since the robot-hand motion in the simulator replays differs from that in the generated videos.
    }
    \label{fig:av_ia_examples}
\end{figure}

\clearpage
\section{Additional Analysis of Attentive Probe}
\label{appendix:attentive_probe_analysis}

We verify the effectiveness and generalization of the attentive probe by evaluating its precision-recall performance on held-out data and under two types of distribution shifts.

\begin{figure}[h]
    \centering
    \caption{
    \textbf{Visualization of distribution shift.} (Left) held-out pair from ActionNet~\citep{fourier2025actionnet}. (Middle) camera-view variation applied to the simulator replay. (Right) visual variation, where the robot video is drawn from a different dataset~\citep{gao2026dreamdojo}.
    }
    \label{figure:probe_analysis_pairs}
    \vspace{4pt}
    \includegraphics[width=0.25\textwidth]{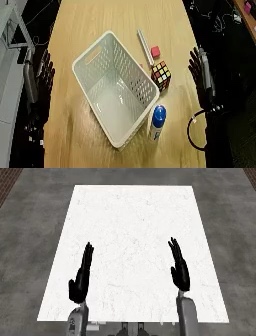}\hspace{2pt}%
    \includegraphics[width=0.25\textwidth]{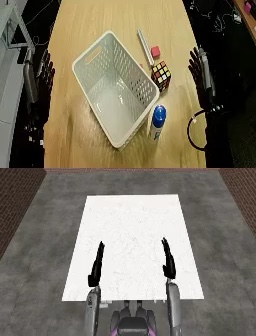}\hspace{2pt}%
    \includegraphics[width=0.25\textwidth]{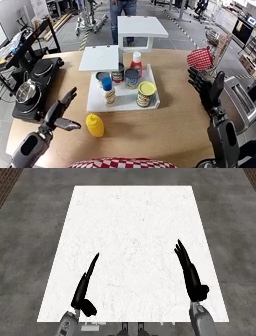}
\end{figure}

\textbf{Experiment setup.} We construct a held-out evaluation set from 100 episodes of ActionNet~\citep{fourier2025actionnet}, drawing positive and negative pairs following the procedure in~\Cref{sec:method} to obtain 500 samples at a 4:1 positive-to-negative ratio. We then introduce two distribution shifts to these samples: (1) visual variation and (2) camera-view variation, which target the video side and the simulator-rollout side of the input pair, respectively. For visual variation, we replace the real video with one from a different robot dataset\footnote{\url{https://huggingface.co/datasets/nvidia/PhysicalAI-Robotics-GR00T-Teleop-GR1}}~\citep{gao2026dreamdojo} that shares the same embodiment. For camera-view variation, we render the RoboCasa~\citep{nasiriany2024robocasa} simulator with the camera offset increased by 0.1\,m and field-of-view widened by 10\textdegree{}. Visualizations of each variation are provided in \Cref{figure:probe_analysis_pairs}.

\textbf{Experiment results.} \Cref{tab:pr_eval} shows that the probe achieves consistent separation between positive and negative samples on held-out ActionNet, with an AUPRC of 0.907. Under camera-view variation, performance remains close to this reference (AUPRC 0.904), suggesting that the probe is not sensitive to small simulator-side changes. Under visual variation, the probe maintains meaningful performance, demonstrating generalization even when the video side is drawn from a different dataset.

\vspace{12pt}

\vspace{12pt}
\begin{table}[h]
\centering
\caption{\textbf{Precision-recall evaluation of the attentive probe across distribution shifts.} We measure AUPRC and precision at fixed recall levels for visual variation (applied to the video side) and camera-view variation (applied to the simulator-rollout side).}
\vspace{0.1in}

\label{tab:pr_eval}
\fontsize{8pt}{9pt}\selectfont
\begin{tabular}{lccccc}
\toprule
\textbf{Setup} & \textbf{AUPRC} & \textbf{P@R=0.25} & \textbf{P@R=0.50} & \textbf{P@R=0.75} & \textbf{P@R=1.00} \\
\cmidrule(lr){1-1} \cmidrule(lr){2-6}
ActionNet~\citep{fourier2025actionnet} (held-out)            & 0.907 & 0.949 & 0.923 & 0.876 & 0.842 \\
Visual variation~\citep{gao2026dreamdojo}    & 0.865 & 0.882 & 0.857 & 0.842 & 0.800 \\
Camera-view variation           & 0.904 & 0.962 & 0.919 & 0.870 & 0.842 \\
\bottomrule
\end{tabular}
\end{table}

\clearpage

\section{Implementation Details on Ablation Studies}
\label{appendix:ablation_details}

\textbf{Filtering strategy ablation details.}  \Cref{tab:ablation_comparison} compares different filtering strategies, including the baselines VideoCon-Physics~\citep{bansal2024videophy} and DreamGenBench~\citep{jang2025dreamgen}.
For VideoCon-Physics, we strictly follow the official implementation, using their VLM fine-tuned for physics alignment evaluation without modification. For DreamGenBench, we strictly follow their exact prompt template using Gemini 2.5 Flash~\citep{comanici2025gemini} with 8 input frames.

\vspace{0.1in}

\section{Inverse Dynamics Models used in \Methodname}
\label{appendix:idm}

IDMs \citep{baker2022video} predict actions between a current observation $\mathbf{o}_t$ and a future observation $\mathbf{o}_{t+H}$, where $H$ denotes the action horizon. Formally, IDMs estimate the intermediate action sequence $\hat{\mathbf{a}}_{t:t+H-1} = \mathrm{IDM}(\mathbf{o}_t, \mathbf{o}_{t+H})$, which is a sequence of actions from timestep $t$ to $t+H-1$.
We use a Diffusion Transformer (DiT) \citep{peebles2023scalable} as our inverse dynamics model, trained with a flow-matching objective~\citep{lipman2022flow}. At inference time, the model denoises the entire action sequence $\hat{\mathbf{a}}_{t:t+H-1}$ conditioned on the input frames. As a result, IDMs serve as a pseudo-action labeler that converts action-free video into action-labeled trajectories.
Concretely, we apply IDMs to synthetic videos generated by an image-to-video (I2V) diffusion model \citep{ali2025world, wan2025wan}, where we chain inferences on frame pairs along the whole video.
This yields synthetic data, consisting of paired synthesized videos and pseudo-labeled action sequences, enabling policy training on action-free data.

\section{More Related Work}
\label{appendix:related_work}
\textbf{Video generative models for robot learning.} Video generative models were originally developed for general human video generation, and prior work has explored using such videos to provide novel scenes or motions for robot policy learning \citep{bharadhwaj2024gen2act}. Another line of work leverages text-to-video {(T2V)} diffusion models to generate synthetic trajectories and {infer} pseudo actions via inverse dynamics models (IDMs) for task generalization \citep{luo2025solving}. More recent studies further explore using video generative models as the robot policy itself, treating video prediction as a proxy for policy learning \citep{wu2023unleashing}. Alongside these trends, \citet{jang2025dreamgen} adopt state-of-the-art video generative models as a data generation pipeline that synthesizes robot videos and extracts pseudo actions for policy training. Our method also follows this data-centric pipeline, using video generation as a source of synthetic experience for learning from limited robot data while explicitly increasing the diversity of manipulation scenes.

\clearpage
\pagebreak
\section{Prompts for VLM}
\label{sec:prompt_for_vlm}

\subsection{Task Instructions for I2V Diffusion Models in Scene Diversification}
\label{appendix:i2v_task_instruction}
\begin{promptbox}{Task instruction generation}
<system>

You are an expert at analyzing robot manipulation scenes and generating diverse, realistic task instructions.

Your task is to generate exactly 5 different plausible robot manipulation tasks that could realistically be performed in the scene.

IMPORTANT: Each instruction must use a SPECIFIC hand as follows:
- Instruction 1: Use the {hands[0]} hand
- Instruction 2: Use the {hands[1]} hand
- Instruction 3: Use the {hands[2]} hand
- Instruction 4: Use the {hands[3]} hand
- Instruction 5: Use the {hands[4]} hand

Each instruction MUST:
1. Follow this exact format: "The robot arm is performing a task. Use the [specified hand] hand to [specific action]."
2. Be based on what you observe in the frame (objects, positions, colors, spatial relationships)
3. Be actionable and realistic for a robot arm
4. Represent a primitive manipulation task doable in 10 seconds
5. Be independent and novel - not variations of each other
6. Use specific object names and descriptive details visible in the scene
7. Have a clear goal condition

<user>
Analyze the image and generate 5 robot manipulation instructions.

IMPORTANT HAND CONSTRAINTS:
- Instruction 1: Must use the {hands[0]} hand
- Instruction 2: Must use the {hands[1]} hand
- Instruction 3: Must use the {hands[2]} hand
- Instruction 4: Must use the {hands[3]} hand
- Instruction 5: Must use the {hands[4]} hand

FOLLOW THESE STEPS:

1. First, let's think step by step through the problem:
   - Identify all visible objects (names, colors, positions)
   - Describe spatial relationships between objects
   - List plausible manipulation tasks that can be done with either hand
   - Plan 5 diverse, novel instructions (each using its specified hand)

2. Generate the response with exactly 5 instructions in the specified format.

REFERENCE EXAMPLES OF INSTRUCTION FORMAT (from similar datasets):
These are examples of high-quality instructions to use as reference:

<example1>
<example2>
<example3>
<example4>
<example5>

Use these examples as style and format reference, but generate COMPLETELY NOVEL instructions based on what you see in the image.

Generate exactly 5 instructions, each following the format:
"The robot arm is performing a task. Use the [specified hand] hand to [action]."

REMEMBER: Each instruction must use its assigned hand from the list above.
Make sure all instructions are diverse and use specific object names from the scene.

\end{promptbox}

\clearpage{}

\subsection{System Prompts for I2I Editing}
\label{appendix:prompts_i2i}
\begin{promptbox}{Initial Scene Description Prompt}
You are given a single image from an indoor tabletop robot manipulation scene. Describe the visible scene in one coherent paragraph from the robot's first-person perspective. Mention the table or main support surface. Mention the main objects on or near the table, their colors, and approximate positions. Mention the surrounding room context briefly (e.g., kitchen, lab, cabinets). Focus only on what is currently visible; do NOT describe actions, motion, or intent. Use natural English in paragraph form (no bullet points)    - INITIAL_SCENE_DESCRIPTION:
\end{promptbox}

\begin{promptbox}{Build Variation Prompt}
You are given:
1) An initial scene description of an indoor tabletop manipulation scenario.
2) A reference instruction describing what interaction happens in the scene.
   Use it as context to understand the type of manipulation, but you are NOT required to keep the same object.

Your goal is to produce structured visual variation candidates for:
[TABLE], [TARGET_OBJECT_VARIANTS], [SENTENCE_LIGHT], and [SENTENCE_BACKGROUND].
All variations must remain visually plausible for an indoor tabletop setup.

IMPORTANT: Generate MULTIPLE diverse candidates for each category to enable varied counterfactual generation. MAXIMIZE DIVERSITY - avoid similar or repetitive descriptions:

- [TABLE]: Generate 2-3 DIFFERENT table/surface descriptions with VARIED patterns, textures, materials, and finishes.
  CRITICAL: Each table description must be DISTINCTLY different from others. Include:
    - Material variety: wood (oak, pine, walnut, bamboo), metal (steel, aluminum, brushed), glass, laminate, marble,...
    - Pattern variety: solid colors, stripes, checkered, wood grain, marble veins, geometric patterns, plain surfaces
    - Texture variety: smooth, rough, matte, glossy, polished, brushed, textured, grainy, satin finish
  Examples of DIVERSE table descriptions:
    - "a dark walnut wooden table with visible grain patterns and a matte finish", -...
  AVOID: Similar descriptions (e.g., don't generate "wooden table" and "oak table" - they're too similar).

- [TARGET_OBJECT_VARIANTS]: Generate 2-3 object variants with varied shapes, colors, and materials.
  CRITICAL: Variants can include BOTH the original target object (with different shape/color/material) AND completely different object types.
  IMPORTANT VARIANT OPTIONS:
    1. Original target object with different attributes: Keep the SAME object TYPE as the reference instruction, but change SHAPE, COLOR, and/or MATERIAL/FINISH.
       Example: If original is "potato", you can generate "a round red potato with a glossy finish" or "a cylindrical brown potato with a matte texture"
    2. Completely different object types: Use DIFFERENT object categories for maximum diversity.
       Example: If original is "potato", you can generate "a cylindrical blue ceramic mug" or "a rectangular green glass bottle"
  CRITICAL - RANDOM COMBINATION FOR EACH VARIANT:
    - For EACH variant, you MUST RANDOMLY combine: TYPE x SHAPE x COLOR x MATERIAL/FINISH
    - TYPE can be: (1) the original target object type, OR (2) a completely different object type
    - If using original target object type: Keep TYPE the same, but RANDOMLY change SHAPE, COLOR, and MATERIAL/FINISH
    - If using different object type: RANDOMLY select TYPE, SHAPE, COLOR, and MATERIAL/FINISH
    - Do NOT use predictable patterns (e.g., don't always use round + red, or cylindrical + blue)
    - RANDOMLY select one type, one shape, one color, and one material/finish for each variant
    - Each variant should have a UNIQUE and RANDOM combination
    - Example of RANDOM combinations (if original is "potato"):
      * Variant 1: potato(type)+round(shape)+red(color)+glossy(finish) = "a round red potato with a glossy finish", * ...
  Examples of HIGHLY DIVERSE object descriptions: - "a round red apple with a glossy finish", -...
  IMPORTANT: Each variant must be a single, natural English phrase that includes: [object type] + [shape] + [color] + [material/finish]. DO NOT include size descriptions (e.g., "small", "large", "medium").
  AVOID: Predictable patterns in type x shape x color combinations.

- [SENTENCE_LIGHT]: Generate 2-3 DIFFERENT lighting descriptions with varied brightness levels and qualities.
  CRITICAL: Only describe the general brightness and lighting quality that illuminates the table surface. 
  IMPORTANT: Maximize diversity across brightness levels and qualities:
    - Brightness levels: bright, dim, soft, harsh, intense, subtle, muted, vibrant, subdued, gentle
    - Lighting qualities: even illumination, diffused light, natural daylight, warm ambient, ...
  Examples of DIVERSE lighting:
    - "bright and even illumination", -...
  DO NOT mention any specific light sources, lamps, fixtures, or objects. DO NOT mention windows, ceiling fixtures, or any physical items. Only describe the overall brightness and lighting quality that affects the table surface.
  AVOID: Similar descriptions (e.g., don't generate "bright light" and "bright illumination" - they're too similar).

- [SENTENCE_BACKGROUND]: Generate 2-3 COMPLETELY DIFFERENT background descriptions with varied environments and styles.
  CRITICAL: Describe ONLY the room environment, walls, and space around/behind the table. You can add distractors on the table OR describe the background environment, but do NOT affect the TABLE or the main target objects.
  IMPORTANT: Maximize diversity across different background types:
    1. Room environments: kitchen, lab, workshop, office, living room, garage, studio, classroom, warehouse, storage room
    2. Wall styles: smooth white walls, textured beige walls, brick walls, painted walls, concrete walls, ...
    3. Distractors on table: fruits/vegetables, tools, books, utensils, random objects (must be non-intrusive)
    4. Background elements: furniture, equipment, shelves, cabinets, windows, doors, artwork, plants, storage items
\end{promptbox}

\begin{promptbox}{Build Variation Prompt - Continued}
  Examples of HIGHLY DIVERSE backgrounds:
    - "sterile lab room with smooth white walls and a faint grid pattern on the floor behind the table", -...
  CRITICAL CONSTRAINTS: - The TABLE description must remain unchanged and unaffected by background descriptions.
    - The distractors or background should NOT interfere with, replace, or modify the TABLE or the main target objects.
    - Distractors should be additional small objects that do NOT touch, overlap, or interfere with the primary manipulation objects.
    - They should only add visual diversity in the background or as non-intrusive elements on the table surface, without affecting the TABLE itself or the target objects.
  AVOID: Similar backgrounds (e.g., don't generate "white walls" and "smooth white walls" - they're too similar).
  
------------------------------------------------------------
INITIAL_SCENE: 
{initial_scene}

REFERENCE_INSTRUCTION (object identity only): 
{reference_instruction}
------------------------------------------------------------

Hard constraints (must follow) - Output MUST contain ALL labeled blocks exactly once.
- Each list item MUST start with "- ", 
- [TARGET_OBJECT_VARIANTS] MUST contain exactly 2-3 complete object descriptions.

General rules
- Indoor tabletop environment only. - Do NOT invent new actions, motion, or outcomes.
- [TARGET_OBJECT_VARIANTS]: Variants can include BOTH the original target object (with different shape/color/material) AND completely different object types. For example, if original is "potato", you can generate: (1) "a round red potato with a glossy finish" (same type, different attributes), (2) "a cylindrical blue ceramic mug" (different type), (3) "a rectangular green glass bottle" (different type). This maximizes diversity while allowing editing
- [TARGET_OBJECT_VARIANTS] - RANDOM COMBINATION: For each variant, RANDOMLY combine TYPE x SHAPE x COLOR x MATERIAL/FINISH. TYPE can be the original target object OR a different object type. Do NOT use predictable or sequential patterns. Each combination must be unique and randomly selected.
- Each variant must be a complete, natural English description. DO NOT include size descriptions.
- [TABLE]: Ensure each table description differs significantly in material, pattern, or texture. Avoid generating multiple wooden tables with only slight variations. RANDOMLY select different material/pattern/texture combinations.
- [SENTENCE_BACKGROUND]: Ensure each background differs significantly in environment type, wall style, or distractor type. Avoid generating multiple "white wall" variations. RANDOMLY select different environment/wall/distractor combinations.
- [SENTENCE_LIGHT]: RANDOMLY select different brightness levels and lighting qualities. Avoid predictable patterns.
- Focus only on scene elements: table, objects, lighting, background.
- Do NOT mention perspective, viewpoint, or robot hands (these will be preserved from the original image).
- Fluent natural English only.
- MAXIMIZE DIVERSITY: When in doubt, choose the more diverse option. The goal is to generate counterfactual scenarios that are visually distinct from each other.
- RANDOM SELECTION: All combinations must be RANDOM - no sequential patterns, no predictable rotations, no fixed orders.

------------------------------------------------------------
Return exactly:

INITIAL_SCENE_REWRITE:
...
[TABLE]:
- ...
[TARGET_OBJECT_VARIANTS]:
- a round red apple with a glossy finish
- a cylindrical blue ceramic mug with a matte texture
- a rectangular green glass bottle with a smooth surface
- a metallic silver wrench with brushed finish
- ...
[SENTENCE_LIGHT]:
- ..
[SENTENCE_BACKGROUND]:
- ...
\end{promptbox}

\begin{promptbox}{Build Counterfactuals Prompt}
You are generating scene-only counterfactual prompts for an image-to-image editing model in STRUCTURED JSON format.
The original image is provided as input to the image-to-image model, so the model will preserve the existing scene structure, robot view, and hands while applying counterfactual changes.
The original image's perspective, viewpoint, and robot hands will be preserved automatically.
You only need to describe the scene elements that should be changed.

You are given:
1) ORIGINAL_SCENE_CAPTION (for context only): {initial_scene_caption}
2) REFERENCE_INSTRUCTION (this has been parsed to extract target object): {reference_instruction}
3) VARIANTS_JSON (candidates): {variants_json} {target_object_note}

------------------------------------------------------------
Your task
Generate exactly {num_instructions} counterfactual prompts in JSON format.
Each prompt must be a valid JSON object following this EXACT structure:

{{
  "scene": "Image 1 is the original scene image. Image 2 is a Canny edge map that defines the strict structural layout and boundaries of all objects in the scene. You MUST strictly follow the edge map structure from Image 2, but ignore the background edges from Image 2 and paint them in as a realistic image. Realistic image of a first-person perspective, from which a robot faces a [table description from TABLE candidates]. All objects on the table must be physically plausible - they must be properly placed on the table surface, not floating in the air, not overlapping with each other, and must follow realistic physics and gravity.",
  "subjects": [
    {{
      "description": "The original {target_object_from_ref if target_object_from_ref else 'target object'} has been edited to [target_object_variant from TARGET_OBJECT_VARIANTS]. The edited object must be a hand-sized object (appropriate size for robot manipulation, not too large or too small). The object must be placed on the table surface, resting naturally without floating or overlapping with other objects. The object must NOT be in the process of being manipulated by the robot - it must be in a static, resting state on the table.",
      "color_palette": ["extract and list colors from the target_object_variant description"]
    }}
  ],
  "style": "Ultra-realistic image. The output must NOT be a Canny edge map - it must be a realistic, photorealistic image with proper textures, colors, shading, and lighting. It must look like a real photograph, NOT like an edge map or line drawing.",
  "lighting": "[ONE lighting sentence from SENTENCE_LIGHT candidates]",
  "background": "[ONE background sentence from SENTENCE_BACKGROUND candidates]"
}}

IMPORTANT - RANDOM COMBINATION RULE (CRITICAL - MUST FOLLOW):
- For each of the {num_instructions} prompts, you MUST randomly select DIFFERENT combinations from each category.
- Do NOT select sequentially (e.g., don't always use the first item from each list).
- Do NOT use any predictable pattern (e.g., don't rotate through items in order).
- For each prompt, RANDOMLY pick one item from [TABLE], one from [TARGET_OBJECT_VARIANTS], one from [SENTENCE_LIGHT], and one from [SENTENCE_BACKGROUND].
- Each prompt should have a UNIQUE combination of all four categories to maximize diversity.
- CRITICAL - TARGET OBJECT RANDOMIZATION:
  - When selecting from [TARGET_OBJECT_VARIANTS], treat each variant as a complete unit and select it randomly.
  - However, if you need to create NEW target object descriptions, you MUST randomly combine:
    * Object TYPE (e.g., apple, mug, bottle, wrench, banana, carrot, book, tool, container, etc.) - RANDOM selection
    * Object SHAPE (e.g., round, cylindrical, rectangular, square, oval, triangular, elongated, ...) - RANDOM selection
    * Object COLOR (e.g., red, blue, green, yellow, orange, black, white, gray, transparent, ...) - RANDOM selection
    * Material/Finish (e.g., ceramic, glass, metal, plastic, wood, rough, polished, brushed, ...) - RANDOM selection
  - The combination of TYPE x SHAPE x COLOR x MATERIAL must be RANDOM for each prompt - do NOT use predictable patterns.
  - Example of RANDOM combinations:
    * Prompt 1: round (shape) + red (color) + apple (type) + glossy (finish) = "a round red apple with a glossy finish"
    * ...
  - Each combination should be UNIQUE and RANDOMLY selected, not following any sequential or predictable pattern.

CRITICAL JSON FORMATTING RULES:
- Each JSON object must be valid JSON (proper quotes, commas, brackets).
- The "scene" field must start with "Image 1 is the original scene image. Image 2 is a Canny edge map that defines the strict structural layout and boundaries of all objects in the scene. You MUST strictly follow the edge map structure from Image 2, but ignore the background edges from Image 2 and paint them in as a realistic image. Realistic image of a first-person perspective, from which a robot faces a [table description]". It MUST also include: "All objects on the table must be physically plausible - they must be properly placed on the table surface, not floating in the air, not overlapping with each other, and must follow realistic physics and gravity."
- The "subjects" field must be an array with exactly one object containing:
  - "description": MUST explicitly state that "The original [target_object] has been edited to [target_object_variant]". It MUST also include: (1) "The edited object must be a hand-sized object (appropriate size for robot manipulation, not too large or too small)", (2) "The object must be placed on the table surface, resting naturally without floating or overlapping with other objects", (3) "The object must NOT be in the process of being manipulated by the robot - it must be in a static, resting state on the table"
  - "color_palette": An array of color strings extracted from the target object variant description
- The "style" field must state that the output must be a realistic, photorealistic image and NOT a Canny edge map
- The "lighting" field must be a simple string describing overall brightness/illumination quality (from [SENTENCE_LIGHT])
- The "background" field must be a simple string describing room environment or distractors (from [SENTENCE_BACKGROUND])
   
\end{promptbox}

\begin{promptbox}{Build Counterfactuals Prompt - Continued}
CRITICAL CONTENT RULES:
- The original target object '{target_object_from_ref if target_object_from_ref else 'target object'}' has been edited to the variant in subjects[0].description
- SCENE PHYSICAL PLAUSIBILITY RULE: All objects on the table must be physically plausible. Objects must be properly placed on the table surface, not floating in the air, not overlapping with each other, and must follow realistic physics and gravity. No objects should appear to be levitating or intersecting with other objects.
- SUBJECTS DESCRIPTION RULE: The "description" field in subjects[0] MUST explicitly state: "The original {target_object_from_ref if target_object_from_ref else 'target object'} has been edited to [variant description]". It MUST also include: (1) The edited object must be a hand-sized object (appropriate size for robot manipulation, not too large or too small), (2) The object must be placed on the table surface, resting naturally without floating or overlapping with other objects, (3) The object must NOT be in the process of being manipulated by the robot - it must be in a static, resting state on the table.
- LIGHTING RULE: Only describe overall brightness and lighting quality. DO NOT mention light sources, lamps, fixtures, windows, or any physical objects.
- BACKGROUND RULE: Either describe room environment/walls/space behind table, OR add distractors. Must NOT affect TABLE or target objects.
- Do NOT mention robot hands, motion, actions, tasks, or intent. - Indoor tabletop scene only.

Return ONLY valid JSON objects, one per line, under [COUNTERFACTUAL_INSTRUCTIONS]:

[COUNTERFACTUAL_INSTRUCTIONS]:
{{
  "scene": "Image 1 is the original scene image. Image 2 is a Canny edge map that defines the strict structural layout and boundaries of all objects in the scene. You MUST strictly follow the edge map structure from Image 2, but ignore the background edges from Image 2 and paint them in as a realistic image. Realistic image of a first-person perspective, from which a robot faces a [table description]. All objects on the table must be physically plausible - they must be properly placed on the table surface, not floating in the air, not overlapping with each other, and must follow realistic physics and gravity.",
  "subjects": [
    {{
      "description": "The original {target_object_from_ref if target_object_from_ref else 'target object'} has been edited to a round red apple with a glossy finish. The edited object must be a hand-sized object (appropriate size for robot manipulation, not too large or too small). The object must be placed on the table surface, resting naturally without floating or overlapping with other objects. The object must NOT be in the process of being manipulated by the robot - it must be in a static, resting state on the table.",
      "color_palette": ["red"]
    }}
  ],
  "style": "Ultra-realistic image. The output must NOT be a Canny edge map - it must be a realistic, photorealistic image with proper textures, colors, shading, and lighting. It must look like a real photograph, NOT like an edge map or line drawing.",
  "lighting": "[ONE lighting sentence from SENTENCE_LIGHT candidates]",
  "background": "[ONE background sentence from SENTENCE_BACKGROUND candidates]"
}}
{{
  "scene": "Image 1 is the original scene image. Image 2 is a Canny edge map that defines the strict structural layout and boundaries of all objects in the scene. You MUST strictly follow the edge map structure from Image 2, but ignore the background edges from Image 2 and paint them in as a realistic image. Realistic image of a first-person perspective, from which a robot faces a [table description]. All objects on the table must be physically plausible - they must be properly placed on the table surface, not floating in the air, not overlapping with each other, and must follow realistic physics and gravity.",
  "subjects": [
    {{
      "description": "The original {target_object_from_ref if target_object_from_ref else 'target object'} has been edited to a cylindrical blue ceramic mug with a matte texture. The edited object must be a hand-sized object (appropriate size for robot manipulation, not too large or too small). The object must be placed on the table surface, resting naturally without floating or overlapping with other objects. The object must NOT be in the process of being manipulated by the robot - it must be in a static, resting state on the table.",
      "color_palette": ["blue"]
    }}
  ],
  "style": "Ultra-realistic image. The output must NOT be a Canny edge map - it must be a realistic, photorealistic image with proper textures, colors, shading, and lighting. It must look like a real photograph, NOT like an edge map or line drawing.",
  "lighting": "[ONE lighting sentence from SENTENCE_LIGHT candidates]",
  "background": "[ONE background sentence from SENTENCE_BACKGROUND candidates]"
}}
... 
\end{promptbox}
\clearpage{}

\subsection{System Prompts for V2V Transfer}
\label{appendix:prompts_v2v}
\begin{promptbox}{Initial Scene Description Prompt}
You are given a single image from an indoor tabletop robot manipulation scene. Describe the visible scene in one coherent paragraph from the robot's first-person perspective. Mention the table or main support surface. Mention the main objects on or near the table, their colors, and approximate positions. Mention the surrounding room context briefly (e.g., kitchen, lab, cabinets). Focus only on what is currently visible; do NOT describe actions, motion, or intent. Use natural English in paragraph form (no bullet points)    - INITIAL_SCENE_DESCRIPTION:
\end{promptbox}

\begin{promptbox}{Build Variation Prompt}
You are given:
1) An initial scene description of an indoor tabletop manipulation scenario.
2) A reference instruction describing what interaction happens in the scene.
Your goal is to produce structured visual variation candidates for:
[TABLE], [TARGET_OBJECT_VARIANTS], [SENTENCE_LIGHT], and [SENTENCE_BACKGROUND].
All variations must remain visually plausible for an indoor tabletop setup.
IMPORTANT: Generate MULTIPLE diverse candidates for each category.

- [TABLE]: Generate 2-3 DIFFERENT table/surface descriptions with VARIED patterns, textures, materials, and finishes.
  CRITICAL: Each table description must be DISTINCTLY different from others. Include:
    - Material variety: wood (oak, pine, walnut, bamboo), metal (steel, aluminum), glass, laminate, marble, ...
    - Pattern variety: solid colors, stripes, checkered, wood grain, marble veins, geometric patterns, ...
    - Texture variety: smooth, rough, matte, glossy, polished, brushed, textured, grainy, satin finish
  Examples of DIVERSE table descriptions:
    - "a dark walnut wooden table with visible grain patterns and a matte finish", - ...
  AVOID: Similar descriptions (e.g., don't generate "wooden table" and "oak table" - they're too similar).
  
- [TARGET_OBJECT_VARIANTS]: Generate 2-3 object variants that STRICTLY PRESERVE the object TYPE and SHAPE from the reference instruction, but vary in COLOR, MATERIAL, and FINISH.
  CRITICAL: You must identify the target object in the reference instruction and keep its TYPE and SHAPE exactly the same.
  WHAT TO KEEP (FIXED):
    - Object Identity (e.g., if it's a "bottle", it must stay a "bottle")
    - Object Shape (e.g., if it's "cylindrical", it must stay "cylindrical")
  WHAT TO VARY (RANDOMIZED):
    - Color (e.g., red, blue, transparent, ...), - Material/Finish (e.g., plastic, glass, brushed, ...)
  CRITICAL - RANDOM COMBINATION FOR EACH VARIANT:
    - For EACH variant, you MUST RANDOMLY combine: COLOR x MATERIAL/FINISH
    - Do NOT use predictable patterns. RANDOMLY select one color and one material/finish for each variant.
    - Example: If original is "potato" (irregular round shape):
      * Variant 1: potato + red (color) + glossy (finish) = "a round red potato with a glossy finish", * ...
  IMPORTANT: Each variant must be a single, natural English phrase that includes: [shape/type] + [color] + [material/finish]. DO NOT include size descriptions.
  
- [SENTENCE_LIGHT]: Generate 2-3 DIFFERENT lighting descriptions with varied brightness levels and qualities.
  CRITICAL: Only describe the general brightness and lighting quality that illuminates the table surface. 
  IMPORTANT: Maximize diversity across brightness levels and qualities:
    - Brightness levels: bright, dim, soft, harsh, intense, subtle, muted, vibrant, subdued, gentle
    - Lighting qualities: even illumination, diffused light, natural daylight, warm ambient, ...
  Examples of DIVERSE lighting: - "bright and even illumination", - ...
  DO NOT mention any specific light sources, lamps, fixtures, or objects. Only describe the overall brightness and lighting.
  
- [SENTENCE_BACKGROUND]: Generate 2-3 COMPLETELY DIFFERENT background descriptions with varied environments and styles.
  CRITICAL: Describe ONLY the room environment, walls, and space around/behind the table.
  IMPORTANT: Maximize diversity across different background types:
    1. Room environments: kitchen, lab, workshop, office, living room, garage, studio, classroom, warehouse, storage room
    2. Wall styles: smooth white walls, textured beige walls, brick walls, painted walls, concrete walls, ...
    3. Distractors on table: fruits/vegetables, tools, containers, utensils, random objects (must be non-intrusive)
  Examples of HIGHLY DIVERSE backgrounds:
    - "sterile lab room with smooth white walls and a faint grid pattern on the floor behind the table", - ...
  CRITICAL CONSTRAINTS: - The TABLE description must remain unchanged. 
    - Distractors must NOT touch or interfere with the primary manipulation objects.
    
INITIAL_SCENE: {initial_scene}
REFERENCE_INSTRUCTION (object identity only): {reference_instruction}
Hard constraints (must follow)
- Output MUST contain ALL labeled blocks exactly once. - Each list item MUST start with "- ".
- [TARGET_OBJECT_VARIANTS] MUST contain exactly 2-3 complete object descriptions.
General rules
- Indoor tabletop environment only. - Do NOT invent new actions, motion, or outcomes.
- [TARGET_OBJECT_VARIANTS]: STRICT CONSTRAINT: The Object TYPE and SHAPE must match the reference instruction exactly.
- [TARGET_OBJECT_VARIANTS] - RANDOM COMBINATION: For each variant, RANDOMLY combine COLOR x MATERIAL/FINISH.
- Each variant must be a complete, natural English description. DO NOT include size descriptions.
- [TABLE]: Ensure each table description differs significantly in material, pattern, or texture.
- [SENTENCE_BACKGROUND]: Ensure each background differs significantly.
- [SENTENCE_LIGHT]: RANDOMLY select different brightness levels and lighting qualities.
- Focus only on scene elements: table, objects, lighting, background.
- Do NOT mention perspective, viewpoint, or robot hands.  - Fluent natural English only.
- RANDOM SELECTION: All combinations must be RANDOM - no sequential patterns.
Return exactly:
INITIAL_SCENE_REWRITE: ..., [TABLE]: ..., [TARGET_OBJECT_VARIANTS]: ..., [SENTENCE_LIGHT]: ..., [SENTENCE_BACKGROUND]: ...
\end{promptbox}
\clearpage{}

\subsection{VLM Pre-filtering for Synthetic Data}
\label{appendix:physical_plausibility}
\begin{promptbox}{Evaluating physical plausibility of video}
Evaluate the robot hand/end-effector for depth- and geometry-consistent motion across three phases. Judge whether the hand's approach/contacts/transport align with the object's apparent depth, size, and shape (including toward/away-from-camera motion).

(1) Move-to-object (approach distance)
Does the hand move a plausible distance to reach the object's depth/position?
No tele-grab: grasp/contact must not occur while the hand is visibly short in depth or with near-zero approach motion.

(2) Grasp (contact precision)
At grasp/contact, is the hand spatially precise relative to the object's size/shape?
Penalize offset grasps, clipping/penetration, or sloppy contact unlikely to stably hold the object.

(3) Move-to-target (transport distance)
After grasping, does the hand (with the object) move a plausible distance/direction toward the target location?
Penalize large object displacement with little hand motion, drifting, or reaching the target without a clear transport trajectory.

Rate plausibility (1-5) with these guidelines:
5 = All three phases look consistent; no noticeable depth/contact/transport issues.
4 = Mostly consistent; at most ONE minor issue (small offset/jitter) that doesn't change the action outcome.
3 = Mixed; ONE clear issue OR multiple minor issues (e.g., slightly short reach, noticeably offset grasp, weakly justified transport).
2 = Largely implausible; ONE major issue or >=2 clear issues (tele-grab, clear mis-grasp, object moves without matching hand motion).
1 = Clearly implausible; >=2 major issues or a severe violation that dominates (obvious tele-grab and/or heavy clipping/drift).
Reply with a single integer (1-5).    
\end{promptbox}

\subsection{Language Re-captioning for Synthetic Data}
\label{appendix:post_processing}
\begin{promptbox}{Detecting wrong instruction}
You are evaluating whether a video correctly follows a given instruction.

Original Instruction: "{instruction}"

The video is shown as a sequence of frames. Please carefully analyze the frames and determine if the video demonstrates the action described in the instruction.

Answer with ONLY "YES" or "NO".

Your answer:
\end{promptbox}

\begin{promptbox}{Recaptioning wrong instruction}
You are generating a natural language instruction that describes the action shown in a video of a robot performing a task.

The video is shown as a sequence of frames. Please carefully analyze the frames and generate a concise, clear instruction that describes what action is being performed.

Here are some examples of the format we're looking for:

{examples_text}

Now, based on the video frames provided, generate a similar instruction that describes the action shown in the video.

Your instruction (one sentence):
\end{promptbox}

\end{document}